\documentclass[preprint,12pt,3p,authoryear]{elsarticle}
\pdfoutput=1
\usepackage{setspace}
\usepackage{lineno}
\usepackage{amsmath, amssymb}
\usepackage{bm}
\usepackage{array,multirow,graphicx}
\usepackage{caption}
\usepackage{hyperref}
\hypersetup{
    colorlinks=true,
    linkcolor=cyan,
    urlcolor=cyan,
}
\usepackage{float}
\usepackage{booktabs}
\usepackage{tikz}
\usetikzlibrary{bayesnet}
\usetikzlibrary{arrows}
\usetikzlibrary{backgrounds}
\usepackage[toc,page]{appendix}

\biboptions{semicolon}
\usepackage[OT1]{fontenc}

\begin{document}
\captionsetup[figure]{labelfont={bf},labelsep=period,name={Fig.}}
\captionsetup[table]{labelfont={bf},labelformat={default}}
\newcommand{\STAB}[1]{\begin{tabular}{@{}c@{}}#1\end{tabular}}

\begin{frontmatter}

\title{A hierarchical Bayesian model to find brain-behaviour associations in incomplete data sets}

\author[label1,label2]{Fabio S. Ferreira\corref{cor1}}
\ead{fabio.ferreira.16@ucl.ac.uk, ferreira.fabio80@gmail.com}
\author[label1,label2]{Agoston Mihalik}
\author[label1,label2,label3]{Rick A. Adams}
\author[label3]{John Ashburner\corref{cor2}}
\author[label1,label2]{Janaina Mourao-Miranda\corref{cor2}}

\address[label1]{Centre for Medical Image Computing, Department of Computer Science, University College London, London, UK}
\address[label2]{Max Planck University College London Centre for Computational Psychiatry and Ageing Research, University College London, UK}
\address[label3]{Wellcome Centre for Human Neuroimaging, University College London, London, UK}

\cortext[cor1]{Corresponding author. 90 High Holborn, Holborn, London WC1V 6LJ.}
\cortext[cor2]{These authors contributed equally to this work.}

\begin{abstract}
Canonical Correlation Analysis (CCA) and its regularised versions have been widely used in the neuroimaging community to uncover multivariate associations between two data modalities (e.g., brain imaging and behaviour). However, these methods have inherent limitations: (1) statistical inferences about the associations are often not robust; (2) the associations within each data modality are not modelled; (3) missing values need to be imputed or removed. Group Factor Analysis (GFA) is a hierarchical model that addresses the first two limitations by providing Bayesian inference and modelling modality-specific associations. Here, we propose an extension of GFA that handles missing data, and highlight that GFA can be used as a predictive model. We applied GFA to synthetic and real data consisting of brain connectivity and non-imaging measures from the Human Connectome Project (HCP). In synthetic data, GFA uncovered the underlying shared and specific factors and predicted correctly the non-observed data modalities in complete and incomplete data sets. In the HCP data, we identified four relevant shared factors, capturing associations between mood, alcohol and drug use, cognition, demographics and psychopathological measures and the default mode, frontoparietal control, dorsal and ventral networks and insula, as well as two factors describing associations within brain connectivity. In addition, GFA predicted a set of non-imaging measures from brain connectivity. These findings were consistent in complete and incomplete data sets, and replicated previous findings in the literature. GFA is a promising tool that can be used to uncover associations between and within multiple data modalities in benchmark datasets (such as, HCP), and easily extended to more complex models to solve more challenging tasks.
\end{abstract}

\end{frontmatter}


\section{Introduction}
\label{intro}
In the past few years, there has been a substantial increase in the application of multivariate methods, such as Canonical Correlation Analysis (CCA) \citep{Hotelling1936}, to identify associations between two data modalities (e.g., brain imaging and behaviour). CCA uncovers underlying associations between two sets of variables by finding linear combinations of variables from each modality that maximise the correlation between them. This is particularly relevant in brain imaging research, where different types of data (e.g., brain structural/functional data, behavioural and cognitive assessments) are collected from the same individuals to investigate the population variability. Moreover, the unsupervised nature of CCA has made it increasingly popular in fields such as psychiatric neuroscience, where there is a lack of objective biomarkers of illness and the diagnostic categories are not reliable \citep{Insel2010, Bzdok2017}. 

CCA and regularised variants of CCA, such as sparse CCA \citep{Waaijenborg2008, LeCao2009, Witten2009}, have been used to identify associations, for instance, between brain connectivity and cognitive/psychopathology measures \citep{Drysdale2016, Xia2018, Mihalik2019}, brain connectivity and general lifestyle, demographic and behavioural measures \citep{Smith2015, Bijsterbosch2018, Lee2019, Li2019, Alnaes2020}, brain structure, demographic and behavioural measures \citep{Monteiro2016, Mihalik2020} and between different brain imaging modalities \citep{Sui2012}.

Nonetheless, these methods have some limitations. First, they do not provide an inherent robust inference approach to infer the relevant associations. This is usually done by assessing the statistical significance of the associations using permutation inference \citep{Winkler2020} or hold-out sets \citep{Monteiro2016, Mihalik2020}. Second, the associations within data modalities, which might explain important variance in the data, are not modelled. Finally, CCA assumes data pairing between data modalities, which is problematic when values are missing in one or both data modalities. This is a common issue in clinical and neuroimaging datasets, in which the missing values usually need to be imputed or removed before applying the models. 

One potential way to address the limitations mentioned above is to solve the CCA problem within a probabilistic framework. \citet{Bach2006} proposed a probabilistic interpretation of CCA, but showed that the maximum likelihood estimates are equivalent to the solution that standard CCA finds. Nevertheless, probabilistic CCA provided an initial step towards robust inference by allowing estimation of the uncertainty of the parameters and it could be used as building block for more complex models, such as Bayesian CCA proposed by \citet{Klami2007} and \citet{Wang2007}. The authors introduced a hierarchical Bayesian extension of CCA by adding suitable prior distributions over the model parameters to automatically infer the number of relevant latent components (i.e., relevant associations) using Bayesian inference. 

Bayesian CCA has some limitations, however: it is not able to uncover associations within data modalities and, in high dimensional data sets, it can be computationally infeasible \citep{Klami2013}. Virtanen and colleagues \citep{Virtanen2011, Klami2013} proposed an extension of Bayesian CCA to solve these two limitations, whilst removing irrelevant latent components (i.e., components that explain little variance). This model was further extended to include more than two data modalities (termed ``groups") and was named Group Factor Analysis (GFA) \citep{Virtanen2012, Klami2015}. Examples of GFA applications are still scarce: it has mostly been used on genomics data \citep{Klami2013, Suvitaival2014, Zhao2016}, drug response data \citep{Khan2014, Klami2015} and task-based fMRI \citep{Virtanen2011, Virtanen2012, Klami2015}. To the best of our knowledge, GFA has not been applied to uncover associations between brain connectivity and behaviour, especially using high dimensional data.

The original GFA implementation still does not addresses the third limitation mentioned above, i.e., it cannot be applied to data modalities with missing data. Therefore, we propose an extension of GFA that can handle missing data and allows more flexible assumptions about noise. We first applied our GFA extension to synthetic data to assess whether it can find known associations among data modalities. We then applied it to data from the Human Connectome Project (HCP) to uncover associations between brain connectivity and non-imaging measures (e.g., demographics, psychometrics and other behavioural measures). We evaluated the consistency of the findings across different experiments with complete and incomplete data sets. Finally, even though the GFA model was proposed for unsupervised tasks, it can also be used as a predictive model: we applied our GFA implementation to synthetic and HCP data to assess whether it was able to predict missing data and non-observed data modalities from those observed, in incomplete data sets.

To illustrate the differences between GFA and CCA, we also applied CCA to both datasets. First, we hypothesised that GFA would replicate previous CCA findings using broadly the same HCP dataset, where previous investigators identified a single mode of population covariation representing a ``positive-negative" factor linking lifestyle, demographic and psychometric measures to specific patterns of brain connectivity \citep{Smith2015}. Second, we expected CCA to show poorer performance when data was missing, whereas GFA results would be consistent across experiments with complete and incomplete data sets. Due to its flexibility and robustness, the proposed GFA extension provides an integrative framework that can be used to uncover associations among multiple data modalities in benchmark neuroimaging datasets. 

\section{Materials and Methods}
We first describe the link between CCA and GFA (Section \ref{methods_ccatogfa}), then we explain how we modified the GFA model to accommodate missing data (Section \ref{methods_ext}) and used it to make predictions (Section \ref{methods_pred}). These subsections are followed by descriptions of experiments where we assess the effectiveness of the model on synthetic data (Section \ref{methods_expsynt}), as well as on HCP data (Section \ref{methods_exphcp}).

\subsection{From CCA to GFA}
\label{methods_ccatogfa}
In this section, we show that the probabilistic extension of CCA serves as a building block for GFA. We begin by describing CCA (Section \ref{methods_cca}), which is followed by the descriptions of probabilistic (Section \ref{methods_pcca}) and Bayesian CCA (Section \ref{methods_bcca}). We finish this section by describing the GFA model and its inference (Section \ref{methods_gfa}).

\subsubsection{CCA}
\label{methods_cca}
Canonical Correlation Analysis was introduced by \citet{Hotelling1936} and is a classical method for seeking maximal correlations between linear combinations of two multivariate data sets, which can be seen as two different data modalities from the same observations or individuals. This can be illustrated using the following notation: $\mathbf{X}^{(1)} \in \mathbb{R}^{D_{1} \times N}$ and $\mathbf{X}^{(2)} \in \mathbb{R}^{D_{2} \times N}$ are two data matrices containing multivariate data from the same \textit{N} observations, where $D_{1}$ and $D_{2}$ denote the number of variables of $\mathbf{X}^{(1)}$ and $\mathbf{X}^{(2)}$, respectively. CCA finds pairs of weight vectors $\mathbf{u}_{k} \in\mathbb{R}^{D_{1} \times 1}$ and $\mathbf{v}_{k} \in\mathbb{R}^{D_{2} \times 1}$ that maximise the correlation between the corresponding projections $\mathbf{u}_{k}^{T}\mathbf{X}^{(1)}$ and $\mathbf{v}_{k}^{T}\mathbf{X}^{(2)}$ (also known as canonical scores), $k=1 ,\dotsc, K$ (where $K$ is the number of canonical directions, also called CCA modes). This is achieved by solving:

\begin{equation} \label{eq1}
\begin{gathered}
	\text{max}_{\mathbf{u}_{k},\mathbf{v}_{k}} \ \mathbf{u}_{k}^{T}\mathbf{X}^{(1)} \mathbf{X}^{(2) T}\mathbf{v}_{k}, \\ 
	\text{ subject to } \ \mathbf{u}_{k}^{T} \mathbf{X}^{(1)}\mathbf{X}^{(1) T} \mathbf{u}_{k} = 1 \ \mbox{and } \mathbf{v}_{k}^{T} \mathbf{X}^{(2)}\mathbf{X}^{(2) T} \mathbf{v}_{k} = 1,
\end{gathered}
\end{equation}
where the variables (i.e., columns of $\mathbf{X}^{(1)}$ and $\mathbf{X}^{(2)}$) are considered to be standardised to zero mean and unit variance. The optimisation problem in Eq. (\ref{eq1}) can be solved using a standard eigenvalue solution \citep{Hotelling1936}, singular value decomposition (SVD) \citep{Uurtio2017}, alternating least squares (ALS) \citep{Golub1994} or non-linear iterative partial least squares (NIPALS) \citep{Wegelin2000}.

As mentioned above, CCA lacks robust inference methods and does not model the associations within data modalities. Probabilistic approaches, such as probabilistic CCA, might be used to overcome these limitations, in which the generative nature of the models offers straightforward extensions to novel models through simple changes of the generative description, and more robust inference methods (e.g., Bayesian inference).     

\subsubsection{Probabilistic CCA}
\label{methods_pcca}
\begin{figure}[H]
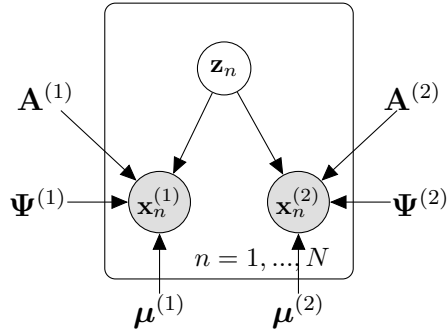

\centering
        \tikz{  
            \node[obs] (x1) {$\mathbf{x}_{n}^{(1)}$};%
            \node[const,left= 0.8 of x1] (s1) {$\boldsymbol{\Psi}^{(1)}$}; %
            \node[const,below= 0.8 of x1] (u1) {$\boldsymbol{\mu}^{(1)}$}; %
            \node[const,above= 0.8 of x1,xshift=-1.5cm] (a1) {$\mathbf{A}^{(1)}$}; %
            \node[obs,right=of x1] (x2) {$\mathbf{x}_{n}^{(2)}$};%
            \node[const,right= 0.8 of x2] (s2) {$\boldsymbol{\Psi}^{(2)}$}; %
            \node[const,below= 0.8 of x2] (u2) {$\boldsymbol{\mu}^{(2)}$}; %
            \node[const,above= 0.8 of x2,xshift=1.5cm] (a2) {$\mathbf{A}^{(2)}$};
            \node[latent,above= of x1,xshift=0.85cm] (z) {$\mathbf{z}_{n}$}; %
            \plate [inner sep=.30cm,yshift=.2cm] {plate1} {(x1)(x2)(z)} {$n=1,...,N$}; %
            \edge {z} {x1,x2} 
            \edge {a1} {x1}
            \edge {u1} {x1}
            \edge {s1} {x1}
            \edge {a2} {x2}
            \edge {u2} {x2}
            \edge {s2} {x2}
            }
\caption{Graphical representation of the probabilistic CCA model.}
\label{fig1}            
\end{figure}

The probabilistic interpretation of CCA \citep{Bach2006} assumes that $N$ observations of $\mathbf{X}^{(1)}$ and $\mathbf{X}^{(2)}$ (similarly defined as above) are generated by the same latent variables $\mathbf{Z} \in\mathbb{R}^{K \times N}$ capturing the associations between data modalities (Fig. \ref{fig1}), where $K$ corresponds to the number of components (which are equivalent to the CCA modes described in Section \ref{methods_cca}): 

\begin{equation} \label{eq2}
\begin{gathered}
	\mathbf{z}_{n}  \sim \mathcal{N}(\mathbf{0}, \mathbf{I}_{K}), \\  
	\mathbf{x}^{(1)}_{n} \sim \mathcal{N}(\mathbf{A}^{(1)}\mathbf{z}_{n} + \boldsymbol{\mu}^{(1)}, \mathbf{\Psi}^{(1)}), \\
    \mathbf{x}^{(2)}_{n} \sim \mathcal{N}(\mathbf{A}^{(2)}\mathbf{z}_{n} + \boldsymbol{\mu}^{(2)}, \mathbf{\Psi}^{(2)}),
\end{gathered}
\end{equation}
where $\mathcal{N}(\cdot)$ represents the multivariate normal distribution,  $\mathbf{A}^{(1)} \in \mathbb{R}^{D_{1} \times K}$ and $\mathbf{A}^{(2)} \in \mathbb{R}^{D_{2} \times K}$ are the projection matrices (also known as loading matrices) that represent the transformations of the latent variables $\mathbf{z}_{n} \in\mathbb{R}^{K \times 1}$ (which corresponds to a column vector of $\mathbf{Z}$) into the input space. The projection matrices are equivalent to the (horizontal) concatenation of all pairs of weight vectors $\mathbf{u}_{k}$ and $\mathbf{v}_{k}$ that CCA finds (see Section \ref{methods_cca}). $\mathbf{\Psi}^{(1)} \in \mathbb{R}^{D_{1} \times D_{1}}$, $\mathbf{\Psi}^{(2)} \in \mathbb{R}^{D_{2} \times D_{2}}$ denote the noise covariance matrices.

Bach and Jordan proved that the maximum likelihood estimates of the parameters in Eq. (\ref{eq2}) lead to the same canonical directions as standard CCA up to a rotation \citep{Bach2006}, i.e., the posterior expectations $E(\mathbf{Z}|\mathbf{X}^{(1)})$ and $E(\mathbf{Z}|\mathbf{X}^{(2)})$ lie in the same subspace that standard CCA finds, where the subspace is represented by the canonical scores $\mathbf{U}^{T}\mathbf{X}^{(1)}$ and $\mathbf{V}^{T}\mathbf{X}^{(2)}$, where $\mathbf{U} \in \mathbb{R}^{D_{1} \times K}$ and $\mathbf{V} \in\mathbb{R}^{D_{2} \times K}$. Moreover, an equivalent representation of the latent variables $\mathbf{Z}$ can be obtained - for CCA - by averaging the canonical scores obtained for each data modality \citep{Klami2013}.

Although probabilistic CCA does not provides an explicit inference approach to infer the number of relevant components, it was used as a building block for Bayesian CCA that - as described in the next section - provides a solution for this limitation. 

\subsubsection{Bayesian CCA}
\label{methods_bcca}
\citet{Klami2007} and \citet{Wang2007} proposed a hierarchical Bayesian extension of CCA by giving full Bayesian treatment to the probabilistic CCA model, introducing suitable prior distributions over the model parameters, which can be inferred using Bayesian inference. 

The goal of Bayesian inference is to provide a procedure for incorporating our prior beliefs with any evidence (i.e., data) that we can collect to obtain an updated posterior belief. This is done using the Bayes' theorem: $p(\mathbf{\Theta}|\mathbf{X}) = p(\mathbf{X}| \mathbf{\Theta})p(\mathbf{\Theta})/p(\mathbf{X})$, where $p(\mathbf{\Theta})$ represents the prior distributions over the model parameters $\mathbf{\Theta}$ (here, $\mathbf{\Theta}$ denotes the model parameters $\{\mathbf{A},\boldsymbol{\alpha},\mathbf{\Psi},\boldsymbol{\mu}\}$ and latent variables $\mathbf{Z}$), $p(\mathbf{X}| \mathbf{\Theta})$ represents the likelihood and $p(\mathbf{\Theta}|\mathbf{X})$ represents the joint posterior distribution that expresses the uncertainty about the model parameters after accounting for the prior knowledge and data. $p(\mathbf{X})$ represents the model evidence, or marginal likelihood, which is usually considered a normalising constant. In this way, Bayes' theorem is formulated as: $p(\mathbf{\Theta}|\mathbf{X}) \propto p(\mathbf{X}| \mathbf{\Theta}) p(\mathbf{\Theta})$, which means that the posterior distribution is proportional to the likelihood times the prior.

\begin{figure}[H]
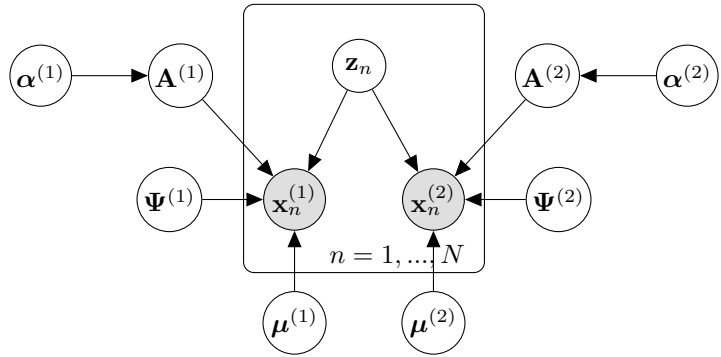

\centering
        \tikz{  
            \node[obs] (x1) {$\mathbf{x}_{n}^{(1)}$};%
            \node[latent,above= of x1,xshift=0.85cm] (z) {$\mathbf{z}_{n}$}; %
            \node[latent,left= 0.8 of x1] (s1) {$\boldsymbol{\Psi}^{(1)}$}; %
            \node[latent,below= 0.8 of x1] (u1) {$\boldsymbol{\mu}^{(1)}$}; %
            \node[latent,above= 0.8 of x1,xshift=-1.5cm] (a1) {$\mathbf{A}^{(1)}$}; %
            \node[latent, left= of a1] (al1) {$\boldsymbol{\alpha}^{(1)}$}; %
            \node[obs,right=of x1] (x2) {$\mathbf{x}_{n}^{(2)}$};%
            \node[latent,right= 0.8 of x2] (s2) {$\boldsymbol{\Psi}^{(2)}$}; %
            \node[latent,below= 0.8 of x2] (u2) {$\boldsymbol{\mu}^{(2)}$}; %
            \node[latent,above= 0.8 of x2, xshift=1.5cm] (a2) {$\mathbf{A}^{(2)}$}; %
            \node[latent,right= of a2] (al2) {$\boldsymbol{\alpha}^{(2)}$}; %
            \plate [inner sep=.25cm,yshift=.2cm] {plate1} {(x1)(x2)(z)} {$n=1,...,N$}; %
            \edge {z} {x1,x2}
            \edge {a1} {x1}
            \edge {u1} {x1}
            \edge {s1} {x1}
            \edge {al1} {a1}
            \edge {a2} {x2}
            \edge {u2} {x2}
            \edge {s2} {x2}
            \edge {al2} {a2}
            }
\caption{Graphical representation of the Bayesian CCA model.}
\label{fig2}            
\end{figure}

In the Bayesian CCA model (represented in Fig. \ref{fig2}), the observations of $\mathbf{X}^{(m)}$ are assumed to be generated by Eq. (\ref{eq2}). The joint probabilistic distribution of the model is given by \citep{Wang2007}:

\begin{equation} \label{eq3} 
\begin{gathered}
	p(\mathbf{X},\mathbf{Z},\mathbf{A},\boldsymbol{\alpha},\mathbf{\Psi},\boldsymbol{\mu}) = \prod_{m=1}^{M} \bigg[ p(\mathbf{X}^{(m)}|\mathbf{Z},\mathbf{A}^{(m)}, \mathbf{\Psi}^{(m)},\boldsymbol{\mu}^{(m)}) \times \\
	p(\mathbf{A}^{(m)}|\boldsymbol{\alpha}^{(m)})p(\boldsymbol{\alpha}^{(m)})p(\mathbf{\Psi}^{(m)})p(\boldsymbol{\mu}^{(m)}) \bigg] p(\mathbf{Z}),
\end{gathered}
\end{equation}
where $M$ is the number of data modalities, $\mathbf{A}^{(m)}$ and $\mathbf{Z}$ are defined as in Eq. (\ref{eq2}) and $\boldsymbol{\alpha}^{(m)} \in\mathbb{R}^{1 \times K}$. The prior distributions are chosen to be conjugate (i.e., the posterior distribution has the same functional form as the prior distribution) which simplifies the inference: 

\begin{equation} \label{eq4}
\begin{gathered}
	p(\mathbf{A}^{(m)}|\boldsymbol{\alpha}^{(m)}) = \prod_{j=1}^{D_{m}} \prod_{k=1}^{K} \mathcal{N}(a_{jk}^{(m)}|0, (\alpha_{k}^{(m)})^{-1}), \quad p(\boldsymbol{\alpha}^{(m)}) = \prod_{k=1}^{K} \Gamma(\alpha_{k}^{(m)}|a_{\boldsymbol{\alpha}}^{(m)},b_{\boldsymbol{\alpha}}^{(m)}), \\
	p(\boldsymbol{\mu}^{(m)}) = \mathcal{N} (\boldsymbol{\mu}^{(m)}|0, (\beta^{(m)})^{-1} \mathbf{I}), \quad p(\mathbf{\Psi}^{(m)}) = \mathcal{W}^{-1}(\mathbf{\Psi}^{(m)}|\mathbf{S}^{(m)}_{0},\nu^{(m)}_{0}),
\end{gathered}
\end{equation}
where $\mathbf{S}_{0}^{(m)}$ is a symmetric positive definite matrix, $\nu^{(m)}_{0}$ denotes the degrees of freedom for the inverse Wishart distribution ($\mathcal{W}^{-1}(\cdot)$) and $\Gamma(\cdot)$ represents the Gamma distribution. The prior over the projection matrices $\mathbf{A}^{(m)}$ is the Automatic Relevance Determination (ARD) prior \citep{Mackay1995}, which is used to find the relevant latent components (i.e., rows of $\mathbf{Z}$). This is done by allowing some $\alpha_{k}^{(m)}$ to be pushed towards infinity, which consequently drives the loadings (i.e., elements of the projection/loading matrices) of the $k$ columns of $\mathbf{A}^{(m)}$ close to zero and the corresponding irrelevant latent components $k$ to be pruned out during inference.   

\sloppy For learning the Bayesian CCA model, we need to infer the model parameters and latent variables from data, which can be done by estimating the posterior distribution $p(\mathbf{Z},\mathbf{A}, \boldsymbol{\alpha}, \mathbf{\Psi},\boldsymbol{\mu}| \mathbf{X})$ and marginalising out uninteresting variables. However, these marginalisations are often analytically intractable and therefore the posterior distribution needs to be approximated. This can be done using mean-field variational Bayes \citep{Wang2007} or Gibbs sampling \citep{Klami2007}, since all conditional distributions are conjugate. However, the inference of the Bayesian CCA model is difficult for high dimensional data as the posterior distribution needs to be estimated over large covariance matrices $\mathbf{\Psi}^{(m)}$ \citep{Klami2013}. The inference algorithms usually need to invert those matrices in every step, which results in $O(D_{m}^{3})$ complexity, leading to long computational times. Moreover, Bayesian CCA does not account for the modality-specific associations.

\citet{Virtanen2011} proposed an extension of Bayesian CCA to impose modality-wise sparsity to separate associations between data modalities from those within data modalities. Moreover, this model assumes spherical noise covariance matrices ($\mathbf{\Psi}^{(m)} = \sigma^{(m)^{2}}\mathbf{I}$, where $\sigma^{(m)^{2}}$ corresponds to the noise variance of data modality $m$) for more efficient inference. The same authors proposed a further extension of the model to uncover associations between more than two groups (e.g., data modalities), called Group Factor Analysis (GFA) \citep{Virtanen2012, Klami2015}.

\subsubsection{Group Factor Analysis}
\label{methods_gfa}
\begin{figure}[H]
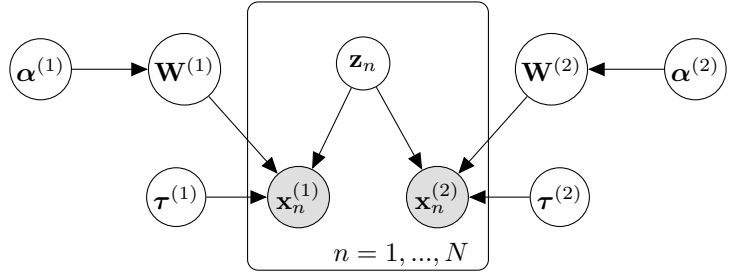

\centering
        \tikz{  
            \node[obs] (x1) {$\mathbf{x}_{n}^{(1)}$};%
            \node[latent,above= of x1,xshift=0.85cm] (z) {$\mathbf{z}_{n}$}; %
            \node[latent,left= 0.8 of x1] (t1) {$\boldsymbol{\tau}^{(1)}$}; %
            \node[latent,above= 0.8 of x1,xshift=-1.5cm] (w1) {$\mathbf{W}^{(1)}$}; %
            \node[latent,left= of w1] (al1) {$\boldsymbol{\alpha}^{(1)}$}; %
            \node[obs,right=of x1] (x2) {$\mathbf{x}_{n}^{(2)}$};%
            \node[latent,right= 0.8 of x2] (t2) {$\boldsymbol{\tau}^{(2)}$}; %
            \node[latent,above= 0.8 of x2, xshift=1.5cm] (w2) {$\mathbf{W}^{(2)}$}; %
            \node[latent,right= of w2] (al2) {$\boldsymbol{\alpha}^{(2)}$}; %
            \plate [inner sep=.25cm,yshift=.2cm] {plate1} {(x1)(x2)(z)} {$n=1,...,N$}; %
            \edge {z} {x1,x2}
            \edge {w1} {x1}
            \edge {t1} {x1}
            \edge {al1} {w1}
            \edge {w2} {x2}
            \edge {t2} {x2}
            \edge {al2} {w2}
            }
\caption{Graphical representation of the GFA model for $M=2$.}
\label{fig3}            
\end{figure}

In the GFA problem, we assume that a collection of $N$ observations, stored in $\mathbf{X} \in \mathbb{R}^{D \times N}$, have disjoint $M$ partitions of variables $D_{m}$ called groups. In this and the following two sections (Sections \ref{methods_ext} and \ref{methods_pred}), we refer to a given data modality as a group of variables of $\mathbf{X}$ ($\mathbf{X}^{(m)} \in \mathbb{R}^{D_{m} \times N}$ for the $m$-th group), in accordance with the GFA nomenclature. Moreover, we introduce the concept ``factor" that corresponds to the loadings in a given column $k$ of the loading matrices (represented as $\mathbf{W}$ in Fig. \ref{fig3}). The latent factors correspond to the rows of the latent variables $\mathbf{Z} \in \mathbb{R}^{K \times N}$ (equivalent to a latent component in probabilistic and Bayesian CCA).

GFA finds the set of $K$ latent factors that can separate the associations between groups (i.e., shared factors) from those within groups (i.e., group-specific factors) by considering a joint factor model (Fig. \ref{fig3}), where each $m$-th group is generated as follows \citep{Virtanen2012, Klami2015}:

\begin{equation} \label{eq5}
\begin{gathered}
    \mathbf{z}_{n} \sim \mathcal{N}(\mathbf{0}, \mathbf{I}_{K}), \\
    \mathbf{x}^{(m)}_{n} \sim \mathcal{N}(\mathbf{W}^{(m)}\mathbf{z}_{n}, \mathbf{T}^{(m)^{-1}}),
\end{gathered}
\end{equation}
where $\mathbf{T}^{(m)^{-1}}$ is a diagonal covariance matrix ($\mathbf{T}^{(m)} = \text{diag}(\boldsymbol{\tau}^{(m)})$, where $\boldsymbol{\tau}^{(m)}$ represents the noise precisions, i.e., inverse noise variances of the $m$-th group), $\mathbf{W}^{(m)} \in \mathbb{R}^{D_{m} \times K}$ is the loading matrix of the $m$-th group and $\mathbf{z}_{n} \in\mathbb{R}^{K \times 1}$ is the latent variable for a given observation $\mathbf{x}^{(m)}_n$ (i.e., column of $\mathbf{X}^{(m)}$). The model assumes zero-mean data without loss of generality. Alternatively, a separate mean parameter could have been included; however, its estimate would converge close to the empirical mean, which can be subtracted from the data before estimating the model parameters \citep{Klami2013}. 

If we consider $M=2$ (also known as Bayesian CCA via group sparsity \citep{Virtanen2011} or Bayesian inter-battery factor analysis \citep{Klami2013}), the noise covariance matrix is given by $\mathbf{T} = \bigl( \begin{smallmatrix} \mathbf{T}^{(1)} & 0 \\ 0 & \mathbf{T}^{(2)} \end{smallmatrix} \bigr)$ and $\mathbf{W} =\begin{bmatrix} \mathbf{A}^{(1)} & \mathbf{B}^{(1)} & \mathbf{0} \\ \mathbf{A}^{(2)} & \mathbf{0} & \mathbf{B}^{(2)}  \end{bmatrix} $, where $\mathbf{A}^{(1)}$ and $\mathbf{A}^{(2)}$ represent the loading matrices containing the shared factors and $\mathbf{B}^{(1)}$ and $\mathbf{B}^{(2)}$ correspond to the loading matrices containing the group-specific factors. The structure of $\mathbf{W}$ and the corresponding latent structure (represented by $\mathbf{Z}$) is learned automatically by imposing group-wise sparsity on the factors, i.e., the matrices $\mathbf{A}$ and $\mathbf{B}$ are not explicitly specified \citep{Klami2013}. This is achieved by assuming independent ARD priors to encourage sparsity over the groups \citep{Virtanen2011, Klami2013}:
\begin{equation} \label{eq6}
    p(\mathbf{W}|\boldsymbol{\alpha}) = \prod_{m=1}^{M} \prod_{j=1}^{D_{m}} \prod_{k=1}^{K} \mathcal{N}(w^{(m)}_{jk}|0, (\alpha_{k}^{(m)})^{-1}), \quad p(\boldsymbol{\alpha}) = \prod_{m=1}^{M} \prod_{k=1}^{K} \Gamma(\alpha^{(m)}_{k}| a_{\boldsymbol{\alpha}^{(m)}},b_{\boldsymbol{\alpha}^{(m)}}),
\end{equation}
which is a simple extension of the single ARD prior used by \citet{Wang2007}. Here, a separate ARD prior is used for each $\mathbf{W}^{(m)}$, which are chosen to be uninformative to enable the automatic pruning of irrelevant latent factors. $\Gamma(\cdot)$ represents a gamma distribution with shape parameter $a_{\boldsymbol{\alpha}^{(m)}}$ and rate parameter $b_{\boldsymbol{\alpha}^{(m)}}$. These separate priors cause groups of variables to be pushed close to zero for some factors $k$ ($\mathbf{w}_{k}^{(m)} \to 0$) by driving the corresponding $\alpha_{k}^{(m)}$ towards infinity. If the loadings of certain factors are pushed towards zero for all groups, the underlying latent factor is deemed inactive and pruned out. \citep{Klami2013}. Finally, the prior distributions over the noise and latent variables $\mathbf{Z}$ are:

\begin{equation} \label{eq7}
    p(\boldsymbol{\tau}) = \prod_{m=1}^{M} \prod_{j=1}^{D_{m}} \Gamma(\tau^{(m)}_{j}|a_{\boldsymbol{\tau}^{(m)}},b_{\boldsymbol{\tau}^{(m)}}), \quad p(\mathbf{Z}) = \prod_{k=1}^{K} \prod_{n=1}^{N} \mathcal{N}(z_{kn}|0,1),
\end{equation}
where $\Gamma(\cdot)$ represents a gamma distribution with shape parameter $a_{\boldsymbol{\tau}^{(m)}}$ and rate parameter $b_{\boldsymbol{\tau}^{(m)}}$. The hyperparameters $a_{\boldsymbol{\alpha}^{(m)}}, b_{\boldsymbol{\alpha}^{(m)}}, a_{\boldsymbol{\tau}^{(m)}}, b_{\boldsymbol{\tau}^{(m)}}$ can be set to a very small number (e.g., $10^{-14}$), resulting in uninformative priors. The joint distribution $p(\mathbf{X,Z,W}, \boldsymbol{\alpha,\tau})$ is hence given by:
\begin{equation} \label{eq8}
    p(\mathbf{X,Z,W},\boldsymbol{\alpha,\tau}) = p(\mathbf{X}|\mathbf{Z,W},\boldsymbol{\tau})p(\mathbf{Z})p(\mathbf{W}|\boldsymbol{\alpha}) p(\boldsymbol{\alpha})p(\boldsymbol{\tau}).
\end{equation}

As mentioned in Section \ref{methods_bcca}, the calculations needed to infer the model parameters and latent variables from data are often analytically intractable. Therefore, the posterior distribution needs to be approximated by applying, for instance, mean field variational approximation (similarly to Bayesian CCA \citep{Wang2007}). This involves approximating the true posterior $p(\boldsymbol{\theta}|\mathbf{X})$ by a suitable factorized distribution $q(\boldsymbol{\theta})$ \citep{Bishop1999}. The marginal log-likelihood ($\ln p(\mathbf{X})$) can be decomposed as follows \citep{Bishop2006}:

\begin{equation} \label{eq9}
\begin{gathered}
    \ln p(\mathbf{X}) = \mathcal{L}(q) + D_{KL}(q||p), \\
    \mathcal{L}(q) = \int q(\boldsymbol{\theta}) \ln \frac{p(\mathbf{X}, \boldsymbol{\theta})}{q(\boldsymbol{\theta})}d\boldsymbol{\theta}, \\
    D_{KL}(q||p) = \int q(\boldsymbol{\theta}) \ln \frac{p(\boldsymbol{\theta}| \mathbf{X})}{q(\boldsymbol{\theta})}d\boldsymbol{\theta},
\end{gathered}
\end{equation}
where $D_{KL}(q||p)$ is the Kullback-Leibler divergence between $q(\boldsymbol{\theta})$ and $p(\boldsymbol{\theta}| \mathbf{X})$ and $\mathcal{L}(q)$ is the lower bound of the marginal log-likelihood. Since $\ln p(\mathbf{X})$ is constant, maximising the lower bound $\mathcal{L}(q)$ is equivalent to minimising the KL divergence $D_{KL}(q||p)$, which means $q(\boldsymbol{\theta})$ can be used to approximate the true posterior distribution $p(\boldsymbol{\theta}|\mathbf{X})$ \citep{Bishop1999}. Assuming that $q(\boldsymbol{\theta})$ can be factorised such that $q(\boldsymbol{\theta}) = \prod_{i} q_{i}(\boldsymbol{\theta}_{i})$, the $\mathcal{L}(q)$ can be maximised with respect to all possible distributions $q_{i}(\boldsymbol{\theta}_{i})$ as follows \citep{Bishop1999, Bishop2006}: 

\begin{equation} \label{eq10}
    \ln q_{i}(\boldsymbol{\theta}_{i}) = \langle \ln p(\mathbf{X, \theta}) \rangle_{j \neq i} + \text{const},
\end{equation}
where $ \langle \cdot \rangle_{j \neq i}$ denotes the expectation taken with respect to $\prod_{j \neq i} q_{j}(\boldsymbol{\theta}_{j})$ for all $j \neq i$. In GFA, the full posterior is approximated by:
\begin{equation} \label{eq11}
    q(\boldsymbol{\theta}) = q(\mathbf{Z}) \prod_{m=1}^{M} \big[ q(\mathbf{W}^{(m)})q(\boldsymbol{\alpha}^{(m)}) q(\boldsymbol{\tau}^{(m)}) \big],
\end{equation}
where $\boldsymbol{\theta}$ denotes the model parameters and latent variables ($\boldsymbol{\theta} = \{\mathbf{Z},\mathbf{W},\boldsymbol{\alpha},\boldsymbol{\tau}\}$). As conjugate priors are used, the free-form optimisation of $q(\boldsymbol{\theta})$ (using Eq. (\ref{eq10})) results in the following analytically tractable distributions:
\begin{equation} \label{eq12}
\begin{gathered}
q(\mathbf{Z}) = \prod_{n=1}^{N} \mathcal{N} (\mathbf{z}_{n}|\boldsymbol{\mu}_{\mathbf{z}_{n}},\mathbf{\Sigma}_{\mathbf{z}_{n}}), \quad q(\mathbf{W}^{(m)}) = \prod_{j=1}^{D_{m}} \mathcal{N}(\mathbf{w}^{(m)}_{j}|\boldsymbol{\mu}_{\mathbf{w}^{(m)}_{j}}, \mathbf{\Sigma}_{\mathbf{w}^{(m)}_{j}}), \\
q(\boldsymbol{\alpha}^{(m)}) = \prod_{k=1}^{K} \Gamma(\alpha^{(m)}_{k}| \Tilde{a}_{\boldsymbol{\alpha}^{(m)}}, \Tilde{b}^{(k)}_{\boldsymbol{\alpha}^{(m)}}), \quad  q(\boldsymbol{\tau}^{(m)}) = \prod_{j=1}^{D_{m}} \Gamma(\tau_{j}^{(m)}| \Tilde{a}^{(j)}_{\boldsymbol{\tau}^{(m)}}, \Tilde{b}^{(j)}_{\boldsymbol{\tau}^{(m)}}),
\end{gathered}
\end{equation} 
where $\mathbf{z}_{n}$ is the $n$-th column of $\mathbf{Z}$ and $\mathbf{w}^{(m)}_{j}$ denotes the $j$-th row of $\mathbf{W}^{(m)}$. The optimisation is done using variational Expectation-Maximization (EM), where the parameters in Eq. (\ref{eq12}) are updated sequentially until convergence, which is achieved when a relative change of the lower bound $\mathcal{L}(q)$ falls below an arbitrary low number (e.g., $10^{-6}$). The recommended choice for the maximal number of latent factors is $K=\operatorname{min}(D_{1}, D_{2})$, but in some settings this leads to large $K$ and consequently long computational times \citep{Klami2013}. In practice, a $K$ value that leads to the removal of some irrelevant latent factors should be a reasonable choice \citep{Klami2013}. In our experiments with synthetic data, we initialised the model with different values of $K$ and the results were consistent across the different experiments (Supplementary Fig. 1).   

\subsection{Our proposed GFA extension}
\label{methods_ext}
To handle missing data, we modified the original GFA model by assuming independent noise for each variable (i.e., diagonal noise) within a group ($p(\boldsymbol{\tau}) = \prod_{m=1}^{M} \prod_{j=1}^{D_{m}} \Gamma( \tau^{(m)}_{j}| a_{\boldsymbol{\tau}^{(m)}}, b_{\boldsymbol{\tau}^{(m)}})$) and modifying the variational update rules similarly to what has been proposed by \citet{Luttinen2010} for variational Bayesian factor analysis. The derivations of the variational EM update rules and lower bound can be found in \ref{App1} and \ref{App2}, respectively. Although we just show here examples of our GFA extension being applied to two data modalities, our Python implementation (Section \ref{methods_code}) can be used for more than two data modalities.   

\subsection{Multi-output and missing data prediction}
\label{methods_pred}
As mentioned above, GFA can be used as a predictive model. As the groups are generated by the same latent variables, the unobserved group of new (test) observations ($\mathbf{X}^{(m)^{\star}}$) can be predicted from the observed ones on the test set ($\mathbf{X}^{-(m)^{\star}}$) using the predictive distribution $p(\mathbf{X}^{(m)^{\star}}| \mathbf{X}^{-(m)^{\star}})$ \citep{Klami2015}. This distribution is analytically intractable but its expectation can be approximated using the parameters learned during the variational approximation (\ref{App1}) as follows \citep{Klami2015}:

\begin{equation} \label{eq13}
\begin{aligned}
    \langle \mathbf{X}^{(m)^{\star}}|\mathbf{X}^{-(m)^{\star}} \rangle = & \langle \mathbf{W}^{(m)}\mathbf{Z} \rangle_{q(\mathbf{W}^{(m)}), \ q(\mathbf{Z}|\mathbf{X}^{-(m)^{\star}})}, \\
        =  & \langle \mathbf{W}^{(m)} \rangle \mathbf{\Sigma}_{Z}^{\star} \langle \mathbf{W}^{-(m)^{T}} \rangle \mathbf{T}^{\star} \mathbf{X}^{-(m)^{\star}},
\end{aligned}
\end{equation} 
where $\langle \cdot \rangle$ denotes expectations, $\mathbf{\Sigma}_{Z}^{\star} = \mathbf{I}_{K} + \sum_{{l \neq m}} \sum_{j}^{D_{l}} \langle \tau_{j}^{(l)}\rangle \langle \mathbf{w}_{j}^{(l)^{T}} \mathbf{w}_{j}^{(l)}\rangle$, $\langle \mathbf{w}_{j}^{(l)^{T}} \mathbf{w}_{j}^{(l)} \rangle = \mathbf{\Sigma}_{\mathbf{w}^{(l)}_{j}} + \boldsymbol{\mu}_{\mathbf{w}_{j}^{(l)}}^{T} \boldsymbol{\mu}_{\mathbf{w}_{j}^{(m)}}$ ($\mathbf{\Sigma}_{\mathbf{w}_{j}^{(m)}}$ and $\boldsymbol{\mu}_{\mathbf{w}_{j}^{(m)}}$ are the variational parameters obtained for $q(\mathbf{W}^{(m)})$ in Equation \ref{eqa11}) and $\mathbf{T}^{\star} = \{ \text{diag} (\langle \boldsymbol{\tau}^{(l)}\rangle)\}_{l \neq m}$. In all experiments, $\langle \mathbf{X}^{(m)^{\star}}|\mathbf{X}^{-(m)^{\star}} \rangle$ was used for prediction. 

Additionally, the missing data can be predicted using Eq. (\ref{eq13}) where, in this case, the observed groups $\mathbf{X}^{-(m)^{\star}}$ correspond to the training observations in group $m$ and the missing data is represented as: $\mathbf{X}^{(m)^{\star}} = \mathbf{X}_{nj \in O_{nj}^{(m)}}^{(m)^{\star}}$, where $O_{nj}^{(m)}$ is the set of indices $(n, j)$ for which the corresponding $x_{nj}^{(m)^{\star}}$ are missing.

\subsection{Experiments}

We begin this section by detailing the experiments that we ran on synthetic data (Section \ref{methods_expsynt}), which is followed by the description of the experiments on the HCP dataset (Section \ref{methods_exphcp}).

\subsubsection{Synthetic data}
\label{methods_expsynt}

We validated the extended GFA model on synthetic data drawn from Eq. (\ref{eq5}). We generated $N = 500$ observations for two data modalities with $D_{1} = 50$ ($\mathbf{X}^{(1)} \in \mathbb{R}^{50 \times 500}$) and $D_{2} = 30$ ($\mathbf{X}^{(2)} \in \mathbb{R}^{30 \times 500}$), respectively. The data modalities were generated from two shared and two modality-specific latent factors, which were manually specified, similarly to the examples generated in \citet{Klami2013} (Fig. \ref{fig4}). The shared factors correspond to latent factor 1 and 2, the latent factor specific to $\mathbf{X}^{(1)}$ is shown in latent factor 4 and the latent factor specific to $\mathbf{X}^{(2)}$ is represented in latent factor 3. The $\boldsymbol{\alpha}^{(m)}$ parameters were set to 1 for the active factors and $10^{6}$ for the inactive ones. The loading matrices $\mathbf{W}^{(m)}$ were drawn from the prior (Eq. (\ref{eq6})) and diagonal noise with fixed precisions ($\tau_{1} = 5 \mathbf{I}_{D_{1}}$ and $\tau_{2} = 10 \mathbf{I}_{D_{2}}$) was added to the observations.

We ran GFA experiments on the following selections of synthetic data:
\begin{enumerate}
    \item \textit{Complete data}.
    \item \textit{Incomplete data}:
    \begin{enumerate}
        \item $20\%$ of the elements of $\mathbf{X}^{(2)}$ were randomly removed.
        \item $20\%$ of the observations (i.e., rows) in $\mathbf{X}^{(1)}$ were randomly removed.
    \end{enumerate}
\end{enumerate}
In all experiments, the model was initialised with $K = 15$ (number of latent factors) to assess whether it can learn the true latent factors while automatically removing the irrelevant ones. In experiment 1, the model was also initialised with $K = 30$ to assess whether the model can still converge to a good solution even if we overestimate the number of latent factors (see Supplementary Fig. 1). 

As the variational approximations for GFA are deterministic and the model converges to a local optimum that depends on the initialisation, all experiments were randomly initialised 10 times. The initialisation with the largest variational lower bound was considered to be the best one. For visualization purposes, we matched the true and inferred latent factors by calculating the maximum similarity (using Pearson's correlation) between them, in all experiments. If a correlation value was negative, the corresponding inferred factor was multiplied by $-1$. The inferred factors with correlations greater than 0.70 were visually compared with the true ones. 

For each random initialisation, in all experiments, the data was split into training (80$\%$) and test (20$\%$) sets. The model performance was assessed by predicting one data modality from the other on the test set (e.g., predict $\mathbf{X}^{(2)}$ from $\mathbf{X}^{(1)}$) using Eq. (\ref{eq13}). The mean and standard deviation of the mean squared error (MSE) (calculated between the true and predicted values of the non-observed data modality on the test set) was calculated across the different initialisations (Table \ref{table1}). The chance level of each experiment was obtained by calculating the MSE between the observations on the test set and the means of the corresponding variables on the training set. 

In the incomplete data experiments, the missing data was predicted using Eq. (\ref{eq13}). We calculated the mean and standard deviation (across initialisations) of the Pearson's correlations between the true and predicted missing values to assess the ability of the model to predict missing data. To compare our results with a common strategy for data imputation in the incomplete data experiments, we ran GFA with complete data, after imputing the missing values using the median of the respective variable. The description of additional experiments with missing data can be found in Supplementary Materials and Methods, including experiments applying CCA to complete and incomplete data sets.

\subsubsection{HCP dataset}
\label{methods_exphcp}
We applied our GFA extension to the publicly available resting-state functional MRI (rs-fMRI) and non-imaging measures (e.g., demographics, psychometrics and other behavioural measures) obtained from 1003 subjects (only these had rs-fMRI data available) of the 1200-subject data release of the HCP (\url{https://www.humanconnectome.org/study/hcp-young-adult/data-releases}). 

In particular, we used the brain connectivity features of the extensively processed rs-fMRI data using pairwise partial correlations between 200 brain regions from a parcellation estimated by independent component analysis. The data processing was identical to \citet{Smith2015}, yielding 19,900 brain variables for each subject (i.e., lower triangular part of the brain connectivity matrix containing pair-wise connectivity among all 200 regions). The vectors were concatenated across subjects to form $\mathbf{X}^{(1)} \in \mathbb{R}^{19900 \times 1001}$. We used 145 items of the non-imaging measures used in \citet{Smith2015} as the remaining measures (SR\_Aggr\_Pct, ASR\_Attn\_Pct, ASR\_Intr\_Pct, ASR\_Rule\_Pct, ASR\_Soma\_Pct, ASR\_Thot\_Pct, ASR\_Witd\_Pct, DSM\_Adh\_Pct, DSM\_Antis\_Pct, DSM\_Anxi\_Pct, DSM\_Avoid\_Pct, DSM\_Depr\_Pct, DSM\_Somp\_Pct) were not available in the 1200-subject data release. The non-imaging matrix contained 145 variables from 1001 subjects ($\mathbf{X}^{(2)} \in \mathbb{R}^{145 \times 1001}$).

Similarly to \citet{Smith2015}, nine confounding variables (acquisition reconstruction software version, summary statistic quantifying average subject head motion during acquisition, weight, height, blood pressure systolic, blood pressure diastolic, hemoglobin A1C measured in blood, the cube-root of total brain and intracranial volumes estimated by FreeSurfer) were regressed out from both data modalities. Finally, each variable was standardised to have zero mean and unit variance. For additional details of the data acquisition and processing, see \citet{Smith2015}.

We ran GFA experiments on the following selections of HCP data:
\begin{enumerate}
    \item \textit{Complete data}.
    \item \textit{Incomplete data}:
    \begin{enumerate}
        \item $20\%$ of the elements of $\mathbf{X}^{(2)}$ were randomly removed.
        \item $20\%$ of the subjects were randomly removed from $\mathbf{X}^{(1)}$.
    \end{enumerate}
\end{enumerate}

In all experiments, the model was initialised with $K = 80$ latent factors. As in the experiments with synthetic data, all experiments were randomly initialised 10 times and the data was randomly split into training ($80\%$) and test ($20\%$) sets. The initialisation with the largest variational lower bound was considered to be the best one.  
The number of factors obtained in all experiments was greater than $60$. Therefore, to facilitate interpretability, we selected the most relevant factors by calculating the relative variance explained (rvar) by each factor $k$ within each data modality $m$ (i.e., $k$-th column of $\mathbf{W}^{(m)}$): 
\begin{equation} \label{eq14}
    \text{rvar}_{k}^{(m)} = \frac{\mathbf{w}^{(m)^{T}}_{k} \mathbf{w}^{(m)}_{k}}{\text{Tr}(\mathbf{W}^{(m)}\mathbf{W}^{(m)^{T}})},
\end{equation}
where $\text{Tr}(\cdot)$ represents the trace of the matrix. The factors explaining more than $7.5\%$ variance within any data modality were considered most relevant. In order to decide whether a given most relevant factor was modality-specific or shared, the ratio between the variance explained (var) by the non-imaging and brain loadings of the $k$-th factor was computed: 
\begin{equation} \label{eq15}
    r_{k} = \frac{\text{var}^{(2)}_{k}}{\text{var}^{(1)}_{k}},
\end{equation}
where $\text{var}_{k}^{(m)} = \frac{\mathbf{w}^{(m)^{T}}_{k} \mathbf{w}^{(m)}_{k}}{\text{Tr}(\mathbf{W}^{(m)}\mathbf{W}^{(m)^{T}} + \mathbf{T}^{(m)^{-1}})}$, and $\mathbf{T}^{(m)^{-1}}$ is the diagonal covariance matrix in Eq. (\ref{eq5}). A factor was consider shared if $0.001 \leq r_{k} \leq 300$, non-imaging specific if $r_{k} > 300 $ or brain-specific if $r_{k} < 0.001$. These values were selected taking into account that there was an imbalance of the total number of variables across data modalities ($\sim$100 times more brain connectivity variables than non-imaging measures). These thresholds were validated in high dimensional synthetic data (Supplementary Table 2). 

To assess whether the missing data affected the estimation of the most relevant factors, we calculated the Pearson's correlations between the factors obtained in the complete data experiment and the incomplete data experiments (Table \ref{table3}).  

In the multi-output prediction task, all non-imaging measures were predicted from brain connectivity on the test set. The model performance was assessed by calculating the mean and standard deviation of the relative MSE (rMSE) between the true and predicted values of each non-imaging measure on the test set, across the different initialisations:
\begin{equation} \label{eq16}
    \text{rMSE}_{j} = \frac{\frac{1}{N} \sum_{n=1}^{N} (x^{(2)}_{nj} - x^{(2)^{*}}_{nj})^{2}}{\frac{1}{N} \sum_{n=1}^{N} (x^{(2)}_{nj})^{2}},
\end{equation}
where $N$ is the number of subjects, $x^{(2)}_{nj}$ and $x^{(2)^{*}}_{nj}$ are the true and predicted non-imaging measure $j$ on the test set. The chance level was obtained by calculating the relative MSE between each non-imaging measure on the test set and the mean of the corresponding non-imaging measure in the training data. 

Similarly to the incomplete data experiments on synthetic data, the missing data was predicted using Eq. (\ref{eq13}) and the mean and standard deviation (across initialisations) of the Pearson's correlations between the true and predicted missing values were calculated.

\subsection{Data and code availability}
\label{methods_code}
\sloppy The data used in this study was downloaded from the Human Connectome Project website (\url{https://www.humanconnectome.org/study/hcp-young-adult/document/extensively-processed-fmri-data-documentation}). All authors involved in data curation and analysis agreed to the HCP open and restricted access data use terms and were granted access. The study was approved by the UCL Research Ethics Committee (Project No. 4356/003). The GFA models and experiments were implemented in Python 3.9.1 and are available here: \url{https://github.com/ferreirafabio80/gfa}. The CCA experiments (Supplementary Materials and Methods) were run in a MATLAB toolkit that will be made publicly available in an open-access platform soon.

\section{Results}
In this section, we present the results of the experiments on synthetic data (Section \ref{res_synt}) and real data from the Human Connectome Project (Section \ref{res_hcp}).

\subsection{Synthetic data}
\label{res_synt}
\begin{figure}[H]
\centering
\includegraphics[width=0.80\textwidth]{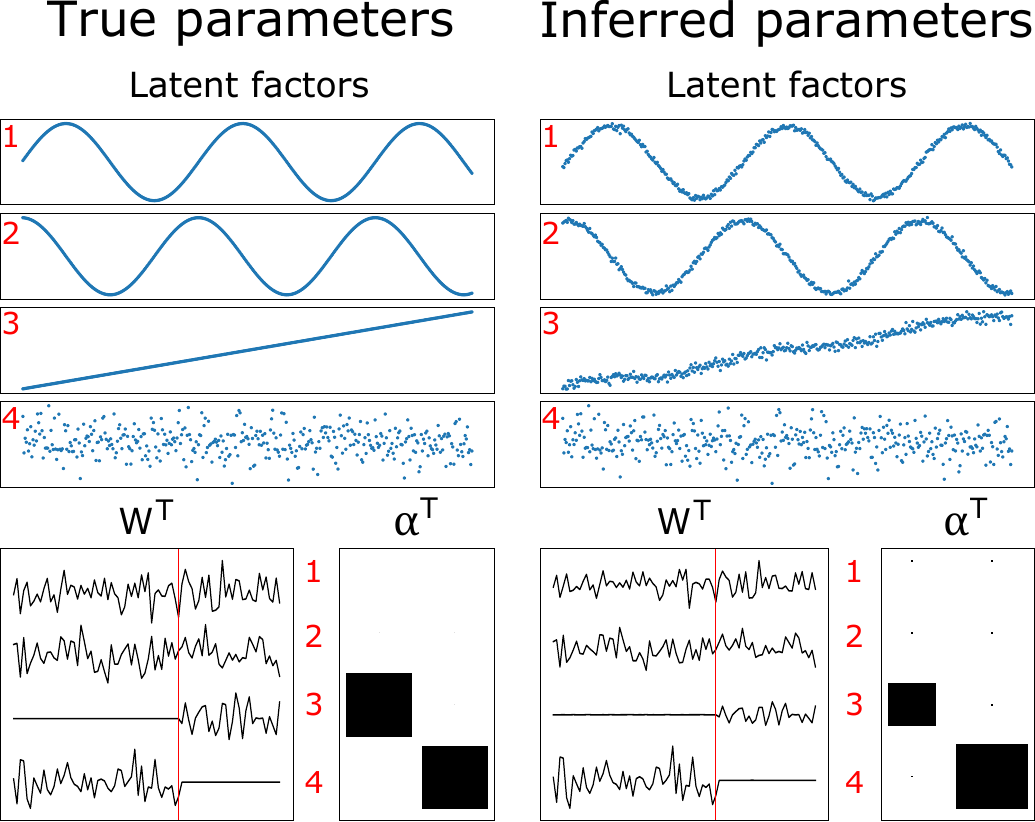}
\caption{True and inferred latent factors and model parameters obtained in the complete data experiment. The latent factors and parameters used to generate the data are plotted on the left-hand side and the ones inferred by the model are plotted on the right-hand side. The four rows on the top represent the four latent factors. The loading matrices of the first and second data modality are represented on the left and right-hand side of the red line in $\mathbf{W}^{T}$, respectively. The alphas of the first and second data modality are shown on the first and second column of $\boldsymbol{\alpha}^{T}$, respectively. The small black dots and big black squares represent active and inactive factors, respectively.}
\label{fig4}
\end{figure}

Fig. \ref{fig4} shows the results of the extended GFA model applied to complete data (experiment 1). The model correctly inferred the factors, identifying two of them as shared and the other two as modality-specific. These factors were all considered most relevant based on the rvar metric (Eq. (\ref{eq14})) and were all correctly assigned as shared or modality-specific based on the ratio $r_{k}$ (Eq. (\ref{eq15})). The structure of the inferred latent factors was similar to those used for generating the data (Fig. \ref{fig4}). The results were robust to initialisation, i.e., the model converged to similar solutions across the different initialisations. Furthermore, the irrelevant latent factors were correctly pruned out during inference. The noise parameters were also inferred correctly (i.e., the average values of $\tau$s were close to the real ones ($\tau_{1} = 5 \mathbf{I}_{D_{1}}$ and $\tau_{2} = 10 \mathbf{I}_{D_{2}}$): $\hat{\tau}^{(1)} \approx 5.08$ and $\hat{\tau}^{(2)} \approx 10.07$).  Furthermore, the model performed well in the multi-output prediction task, i.e., the averaged MSE was lower than chance level when predicting $\mathbf{X}^{(1)}$ from $\mathbf{X}^{(2)}$, and vice-versa (Table \ref{table1}). 

\begin{table}[H]
    \scriptsize
	\centering
	\renewcommand{\tabcolsep}{3pt}
    \caption{Prediction errors of the multi-output prediction tasks. The values correspond to the mean and standard deviation of the MSEs across 10 initialisations. The first row shows the MSE between the test observations $\mathbf{X}^{(1)^{\star}}$ and the mean predictions $\langle \mathbf{X}^{(1)^{\star}}|\mathbf{X}^{(2)^{\star}} \rangle$ of all experiments. On the second row, the MSEs between $\mathbf{X}^{(2)^{\star}}$ and $\langle \mathbf{X}^{(2)^{\star}}|\mathbf{X}^{(1)^{\star}} \rangle$ are shown. ours - proposed GFA approach; imputation - median imputation approach; chance - chance level. Experiment 1 - complete data; experiment 2a - $20\%$ of the elements of $\mathbf{X}^{(2)}$ missing; experiment 2b - $20\%$ of the rows of $\mathbf{X}^{(1)}$ missing.}
    \label{table1}
    \begin{tabular*}{\textwidth}{c|cc|ccc|ccc}
    \toprule
      {} & \multicolumn{2}{c|}{\textbf{Experiment 1}} &  \multicolumn{3}{c|}{\textbf{Experiment 2a}} & \multicolumn{3}{c}{\textbf{Experiment 2b}} \\
      {} & ours & chance & ours & imputation & chance & ours & imputation & chance\cr 
    \midrule
      $\mathbf{X}^{(1)}$ from $\mathbf{X}^{(2)}$ & 1.38 $\pm$ 0.21 & 2.48 $\pm$ 0.28 & 1.23 $\pm$ 0.25 & 1.27 $\pm$ 0.25 & 2.29 $\pm$ 0.27 & 1.14 $\pm$ 0.19 & 1.17 $\pm$ 0.18 & 2.27 $\pm$ 0.26 \\
      $\mathbf{X}^{(2)}$ from $\mathbf{X}^{(1)}$ & 0.81 $\pm$ 0.18 & 2.24 $\pm$ 0.39 & 0.71 $\pm$ 0.11 & 0.74 $\pm$ 0.11 & 2.06 $\pm$ 0.29 & 0.75 $\pm$ 0.18 & 0.75 $\pm$ 0.18 & 2.22 $\pm$ 0.36 \\
     \bottomrule
    \end{tabular*}
\end{table}

\begin{figure}[H]
\centering
\includegraphics[width=\textwidth]{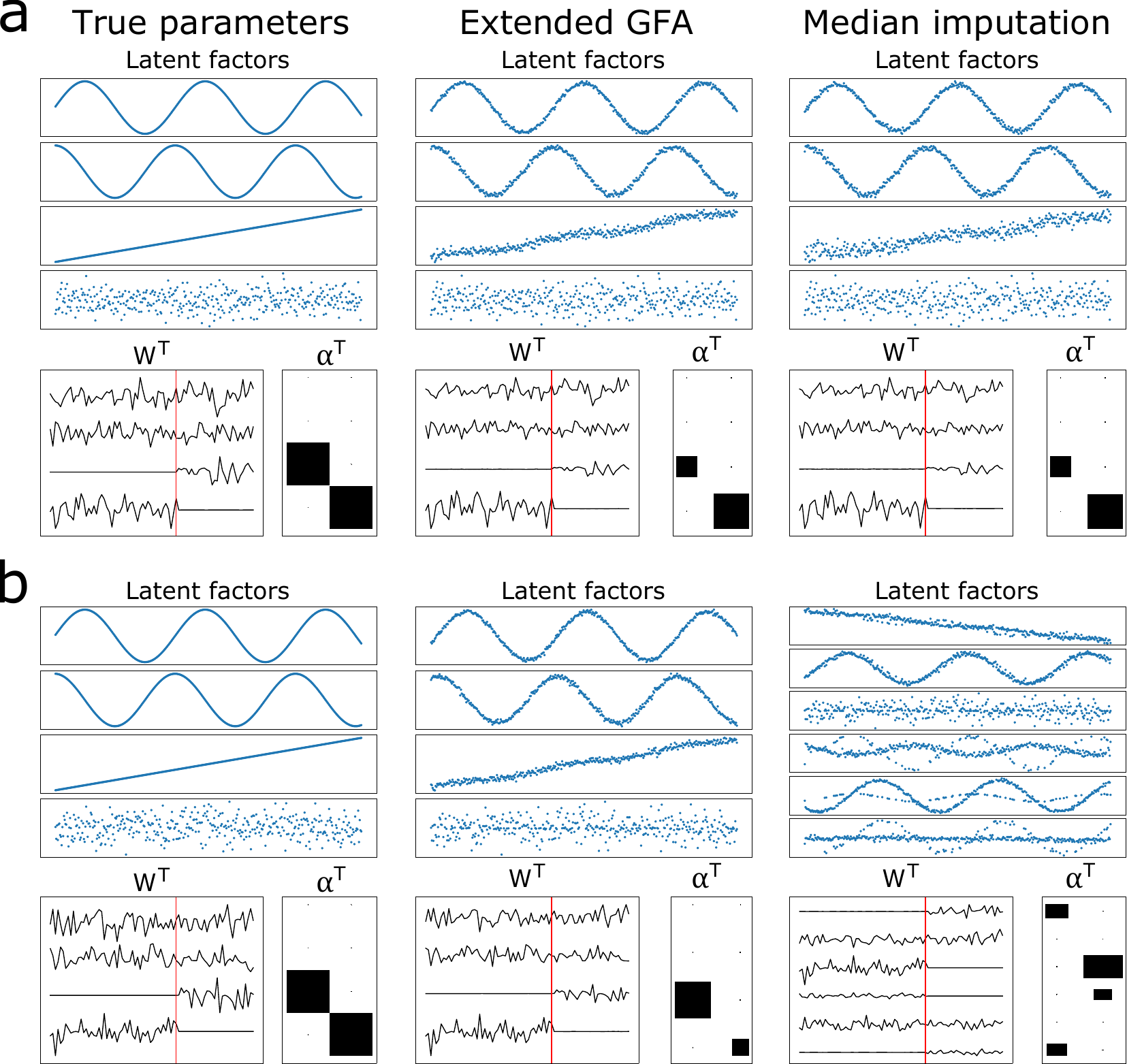}
\caption{True and inferred latent factors and model parameters obtained in the incomplete data experiments 2a \textbf{(a)} and 2b \textbf{(b)}. \textbf{(Left column)} latent factors and parameters used to generate the data. \textbf{(Middle column)} latent factors and parameters inferred using the proposed GFA extension. \textbf{(Right column)} latent factors and parameters inferred using the median imputation approach (the latent factors were not ordered because the model did not converge to the right solution). The loading matrices ($\mathbf{W}^{T}$) and alphas ($\boldsymbol{\alpha}^{T}$) can be interpreted as in Fig. \ref{fig4}.}
\label{fig5}
\end{figure}

Fig. \ref{fig5}a and \ref{fig5}b display the results of the incomplete data experiments 2a ($20\%$ of the elements of $\mathbf{X}^{(2)}$ missing) and 2b ($20\%$ of the rows of $\mathbf{X}^{(1)}$ missing), respectively. The parameters inferred using our GFA extension (Fig. \ref{fig5}, middle column) were compared to those obtained using the median imputation approach (right column). The results were comparable when the amount of missing data was small (Fig. \ref{fig5}a), i.e., both approaches were able to infer the model parameters fairly well. Even so, the model misses completely the true value of the noise parameter of $\mathbf{X}^{(2)}$ ($\hat{\tau}^{(1)} \approx 5.14$ and $\hat{\tau}^{(2)} \approx 5.22$) when the median imputation approach is used. Whereas the noise parameters were correctly recovered ($\hat{\tau}^{(1)} \approx 5.15$ and $\hat{\tau}^{(2)} \approx 10.17$) when the proposed GFA approach was applied. The parameters were not inferred correctly by the median imputation approach (although the noise parameters were recovered fairly well, $\hat{\tau}^{(1)} \approx 6.24$ and $\hat{\tau}^{(2)} \approx 10.20$), when the number of missing observations was considerable (Fig. \ref{fig5}b). This was not observed when our GFA extension was applied ($\hat{\tau}^{(1)} \approx 5.04$ and $\hat{\tau}^{(2)} \approx 10.24$). 

The extended GFA model predicted missing data consistently well in both experiments: $\rho = 0.868 \pm 0.016$ and $\rho = 0.680 \pm 0.039$ (where $\rho$ represents the averaged Pearson's correlation between the missing and predicted values across initialisations) for the incomplete data experiments 2a and 2b, respectively.

In the multi-output prediction task, we showed that the model could make reasonable predictions when the data was missing randomly or one modality was missing for some observations, i.e., the MSEs were similar across experiments and below chance level (Table \ref{table1}). Moreover, there seems to be no improvement between using the proposed GFA approach or imputing the median before training the model.  

In additional experiments (presented in the Supplementary Materials and Methods), we showed that the proposed GFA approach outperforms the median imputation approach (in inferring the model parameters and predicting one unobserved data modality from the other), when values from the tails of the data distribution are missing (Supplementary Fig. 2a and Supplementary Table 1). The proposed GFA approach also outperformed the median imputation approach, when both data modalities were generated with missing values in low (Supplementary Fig. 2b) and high dimensional (Supplementary Fig. 2c) data.

\subsection{HCP data}
\label{res_hcp}
\begin{figure}[H]
\centering
\includegraphics[width=0.75\linewidth]{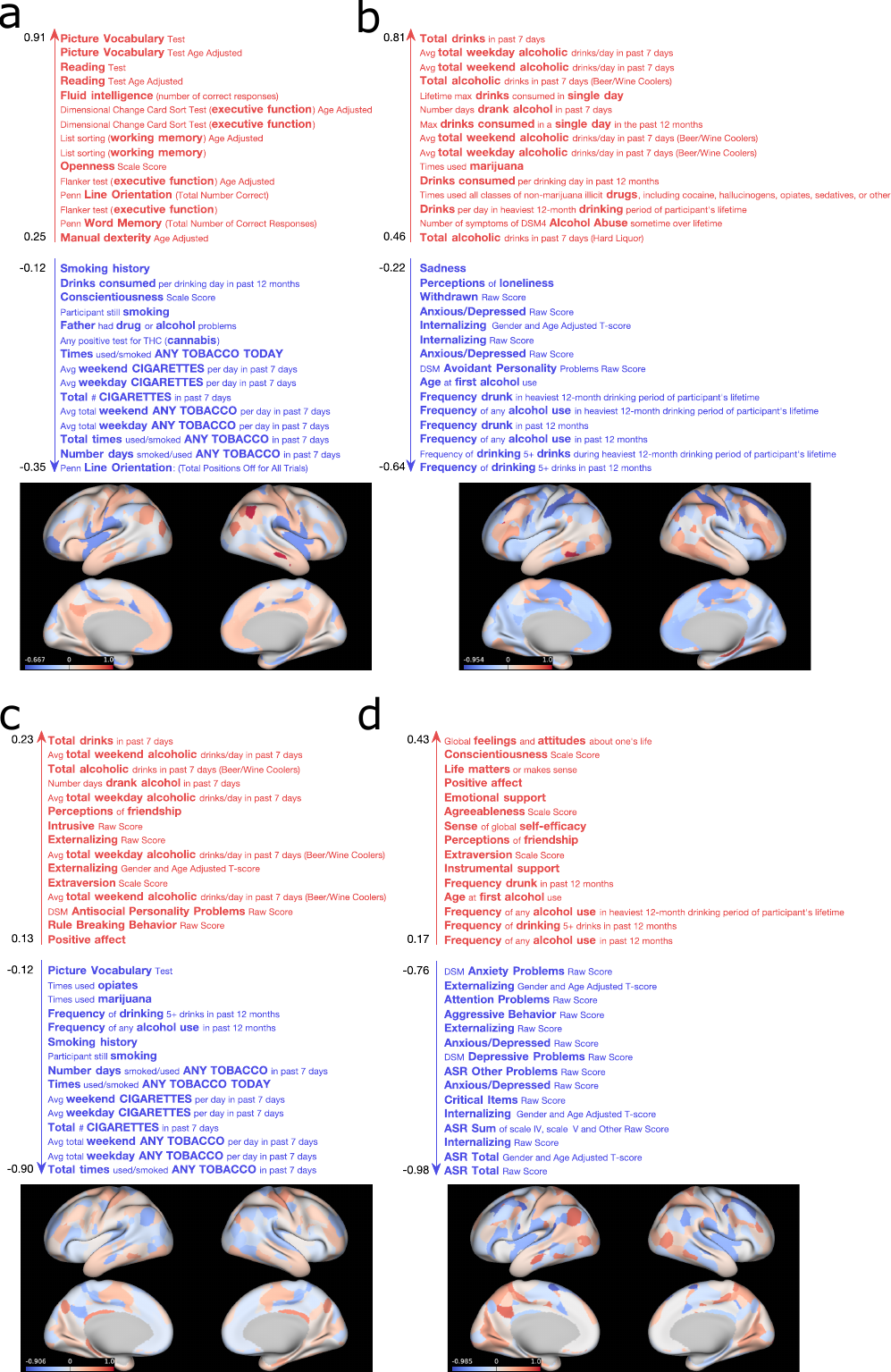}
\caption{Non-imaging measures and brain networks described by the first \textbf{(a)}, second \textbf{(b)}, third \textbf{(c)} and fourth \textbf{(d)} shared GFA factors obtained in the complete data experiment. For illustrative purposes, the top and bottom 15 non-imaging measures of each factor are shown. The brain surface plots represent maps of brain connection strength increases/decreases, which were obtained by weighting each node’s parcel map with the GFA edge-strengths summed across the edges connected to the node (for details, see the Supplementary Materials and Methods). Separate thresholded maps of brain connection strength increases and decreases can be found in Supplementary Fig. 7.}
\label{fig6}
\end{figure}

\begin{figure}[H]
\centering
\includegraphics[width=\linewidth]{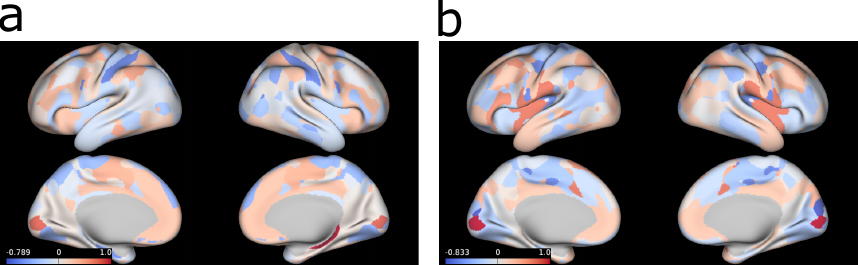}
\caption{Brain networks associated with the brain-specific GFA factors obtained in the complete data experiment. The brain surface plots represent maps of brain connection strength increases/decreases, which were obtained by weighting each node’s parcel map with the GFA edge-strengths summed across the edges connected to the node (for details, see the Supplementary Materials and Methods).}
\label{fig7}
\end{figure}

In the complete data experiment, the model converged to a solution comprising 75 latent factors, i.e., five factors were inactive for both data modalities (the loadings were close to zero) and were consequently pruned out. The model converged to similar solutions across different initialisations, i.e., the number of inferred latent factors was consistent across initialisations. The total percentage of variance explained by the latent factors ($\sum_{m=1}^{2} \sum_{k=1}^{75} \text{var}_{k}^{(m)}$)  corresponded to $\sim$7.55$\%$, leaving $92.45\%$ of the variance captured by residual error. Within the variance explained, six factors were considered most relevant ($\text{rvar}_{k}^{(m)} > 7.5\%$), which captured $\sim$27.8$\%$ of the variance explained by the total number of factors (Table \ref{table2}). Based on the ratio between the variance explained by the non-imaging and brain factors $r_{k}$ (Eq. (\ref{eq15})), we identified four shared factors (displayed in Fig. \ref{fig6}) and two brain-specific factors (displayed in Fig. \ref{fig7}), ordered from the highest to lowest ratio $r_{k}$ (Table \ref{table2}). 

In Fig. \ref{fig6}, we display the loadings of the shared GFA factors obtained with complete data. To aid interpretation, the loadings of the brain factors were multiplied by the sign of the population mean correlation to obtain a measure of edge strength increase or decrease (as in \citet{Smith2015}). The first factor (Fig. \ref{fig6}a) relates cognitive performance (loading positively), smoking and drug use (loading negatively) to the default mode and frontoparietal control networks (loading positively) and insula (loading negatively). The second shared factor (Fig. \ref{fig6}b) relates negative mood, long term frequency of alcohol use (loading negatively) and short term alcohol consumption (loading positively) to the default mode and dorsal and ventral attentional networks (loading negatively), and frontoparietal networks loading in the opposite direction. The third shared factor (Fig. \ref{fig6}c) is dominated by smoking behaviour (loading negatively) and, with much lower loadings, externalising in the opposite direction, which are related to the somatomotor and frontotemporal networks (loading positively). The fourth shared factor (Fig. \ref{fig6}d) seems to relate emotional functioning, with strong negative loadings on a variety of psychopathological aspects (including both internalising and externalising symptoms), and positive loadings on traits such as conscientiousness and agreeableness and other aspects of wellbeing to cingulo-opercular network (loading negatively), and the left sided default mode network (loading positively).

\begin{table}[H]
    \scriptsize
	\centering
	\renewcommand{\tabcolsep}{16pt}
    \caption{Most relevant shared and modality-specific factors obtained with complete data according to the proposed criteria. Factors explaining more than $7.5\%$ variance within any data modality were considered most relevant. A factor was considered shared if $0.001 \leq r_{k} \leq 300$, non-imaging (NI) specific if $r_{k} > 300 $ or brain-specific if $r_{k} < 0.001$. rvar - relative variance explained; var - variance explained; $r_{k}$ - ratio between the variance explained by the non-imaging and brain loadings in factor $k$.}
    \label{table2}
    \begin{tabular}{cc|cc|cc|c}
    \toprule
      {} & {} & \multicolumn{2}{c|}{\textbf{rvar} ($\%$)} & \multicolumn{2}{c|}{\textbf{var} ($\%$)} & $\mathbf{r_{k}}$\\
      \multicolumn{2}{c|}{Factors} & Brain & NI & Brain & NI & var$_{\text{NI}}$/var$_{\text{brain}}$\cr 
    \midrule
      \multirow{4}{*}{\STAB{\rotatebox[origin=c]{90}{\textbf{Shared}}}} & a & 0.096 & 8.103 & 0.007 & 0.028 & 4.03 \\
      & b & 0.032 & 17.627 & 0.002 & 0.061 & 26.22   \\
      & c & 0.011 & 9.869 & 7.65 $\times$ 10$^{-4}$ & 0.034 & 44.32  \\
      & d & 0.008 & 33.336 & 5.46 $\times$ 10$^{-4}$ & 0.114 & 209.65  \\
    \midrule
    \multirow{2}{*}{\STAB{\rotatebox[origin=c]{90}{\textbf{Brain}}}} & a & 14.267 & 2.311 $\times$ 10$^{-9}$ & 1.028 & 7.93 $\times$ 10$^{-12}$ & 7.72 $\times$ 10$^{-12}$  \\
      & b & 11.407 & 0.036 & 0.822 & 1.23 $\times$ 10$^{-4}$ & 1.50 $\times$ 10$^{-4}$  \\
    \bottomrule
    \end{tabular}
\end{table}

Fig. \ref{fig7} shows the loadings of the brain-specific factors obtained with complete data. The first factor (Fig. \ref{fig7}a) contains positive loadings on many areas within the frontoparietal control network, including dorsolateral prefrontal areas and inferior frontal gyrus, supramarginal gyrus, posterior inferior temporal lobe and parts of the cingulate and superior frontal gyrus. The second factor (Fig. \ref{fig7}b) includes positive loadings on many default mode network areas, such as medial prefrontal, posterior cingulate and lateral temporal cortices, and parts of angular and inferior frontal gyri. These factors show that there is great variability in these networks across the sample, however this variability was not linked to the non-imaging measures included in the model.

The model converged to a similar solution in the incomplete data experiment 2a ($20\%$ of the elements of the non-imaging matrix missing), which included 73 factors and the total percentage of variance explained by these was $\sim$7.60$\%$. The number of most relevant factors, based on the rvar metric (Eq. (\ref{eq14})), was six and they were similar to those obtained in the complete data experiment (Table \ref{table3}), capturing $\sim$28.2$\%$ of the variance explained by all factors (Supplementary Table 3). Four of these were considered shared factors (Supplementary Fig. 4) and two were considered brain-specific (Supplementary Fig. 6a,c). In the incomplete data experiment 2b ($20\%$ of the subjects missing in the brain connectivity matrix), the model converged to a solution containing 63 factors and the total percentage of variance explained corresponded to $\sim$5.21$\%$. Although more factors were removed and a loss of variance explained was noticeable, the most relevant factors were similar to those obtained in the other experiments (Table \ref{table3}, Supplementary Fig. 5 and Supplementary Fig. 6b,d), capturing $\sim$33.2$\%$ of the variance explained by all factors (Supplementary Table 4).    

\begin{table}[H]
    \scriptsize
	\centering
	\renewcommand{\tabcolsep}{3pt}
    \caption{Similarity (measured by Pearson’s correlation) between the most relevant factors obtained in the complete and the most relevant factors obtained in the incomplete data experiment 2a and 2b (first and second row, respectively). The shared factors obtained with complete data are displayed in Fig. \ref{fig6} and those obtained in experiment 2a and 2b are shown in Supplementary Fig. 4-5. The brain-specific factors obtained with complete data are presented in Fig. \ref{fig7} and those identified in experiment 2a and 2b are shown in Supplementary Fig. 6.}
    \label{table3}
    \begin{tabular*}{\textwidth}{@{\extracolsep{\fill}}c|cccc|cc}
    \toprule
      {} & \multicolumn{4}{c|}{\textbf{Shared factors}} & \multicolumn{2}{c}{\textbf{Brain factors}}\\
      {} & a & b & c & d & a & b\cr 
    \midrule
      Experiment 2a & 0.896 & 0.964 & 0.954 & 0.989 & 0.974 & 0.974 \\
      Experiment 2b & 0.907 & 0.973 & 0.954 & 0.995 & 0.941 & 0.942  \\
    \bottomrule
    \end{tabular*}
\end{table}

\begin{figure}[H]
\centering
\includegraphics[width=\linewidth]{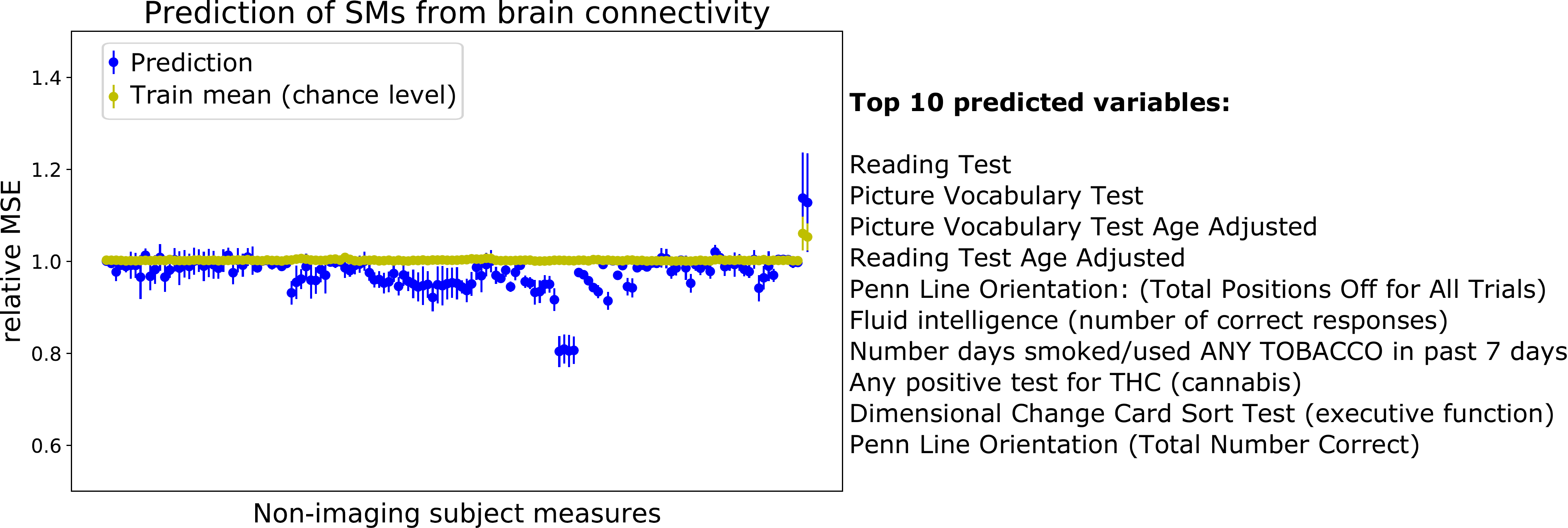}
\caption{Multi-output predictions of the non-imaging measures using complete data. The top 10 predicted variables are described on the right-hand side. For each non-imaging measure, the mean and standard deviation of the relative MSE (Eq. (\ref{eq16})) between the true and predicted values on the test set was calculated across different random initialisations of the experiments.}
\label{fig8}
\end{figure}

In the multi-output prediction task, the proposed GFA model predicted several non-imaging measures better than chance (Fig. \ref{fig8}) using complete data (experiment 1). The top 10 predicted variables corresponded to those with the highest loadings obtained mainly in the first shared factor (Fig. \ref{fig6}a) and were consistent across the incomplete data experiments (Supplementary Fig. 8). Finally, our GFA extension failed to predict the missing values in both incomplete data experiments: $\rho = 0.112 \pm 0.011$ for experiment 2a; $\rho = 0.003 \pm 0.007$ for experiment 2b. 

\section{Discussion}
In this study, we proposed an extension of the Group Factor Analysis (GFA) model that can uncover multivariate associations among multiple data modalities, even when these modalities have missing data. We showed that our proposed GFA extension can: (1) find associations between high dimensional brain connectivity data and non-imaging measures (e.g., demographics, psychometrics and other behavioural measures) and (2) predict non-imaging measures from brain connectivity when either data is missing at random or one modality is missing for some subjects. Moreover, we replicated previous findings obtained in a subset of the HCP dataset using CCA \citep{Smith2015}.

We showed, using synthetic data, that our GFA extension can correctly learn the underlying latent structure, i.e., it separates the shared factors from the modality-specific ones, when data is missing. Moreover, it inferred the model parameters better than the median imputation approach in different incomplete data scenarios. Whereas CCA was only able to recover the shared latent factors and identified spurious latent factors when the values of the tails of the data distribution were missing (Supplementary Fig. 3). These findings underline the importance of using approaches that can handle missing data and model the modality-specific associations. Interestingly, in the multi-output prediction task, our GFA extension only outperformed the median imputation approach when the most informative values of the data (i.e., the values on the tails of the data distribution) were missing (Supplementary Table 1). This indicates that these values might be driving the predictions and the model fails to predict one data modality from the other when these values are not carefully imputed. Finally, our GFA extension was able to predict the missing values in different incomplete data scenarios.

In applying our GFA extension to the HCP dataset, we identified six most relevant factors: four describing associations between brain connectivity and non-imaging measures and two describing associations within brain connectivity. Importantly, these were consistent across the experiments with complete and incomplete data sets. Of note, only a small proportion of the variance was captured by the GFA latent structure, which may be explained by two main reasons: the brain connectivity data is noisy and/or the shared variance between the included non-imaging measures and the brain connectivity measures is relatively small with respect to the overall variance in brain connectivity. Interestingly, most of the featured domains of non-imaging measures were not unique to particular factors, but appeared in different arrangements across the four factors. For instance, alcohol use appeared in three out of four factors: in the first, it loads in the opposite direction to cognitive performance, in the second, its frequency loads in the same direction as low mood and internalising, and in the third, its total amount loads in the same direction as externalising. For a more detailed discussion about the alcohol use loadings, see Supplementary Results.

The first GFA factor was almost identical to the first CCA mode (Supplementary Fig. 9 and Supplementary Table 5), which resembled the CCA mode obtained using a subset of this data set \citep{Smith2015}. The second and third CCA modes presented similar most positive and negative non-imaging measures to the first GFA factor (for a more detailed description of the CCA modes, see the Supplementary Results). A possible explanation of the differences observed between the CCA and GFA results is that we had to apply principal component analysis to reduce the dimensionality of the data before applying CCA. This extra preprocessing step makes the CCA approach less flexible because the model cannot explore all variance in the data, whereas in GFA this does not happen because no dimensionality reduction technique is needed. For more details about the HCP experiments using CCA, see Supplementary Materials and Methods. 

The brain-specific factors were difficult to interpret - as would be expected due to the inherent complexity of this data modality. Their partial similarity to known functional connectivity networks (frontoparietal and default mode) indicates, unsurprisingly, that there are aspects of these networks that are not related to the non-imaging measures included here. Interestingly, the second brain factor (Fig. \ref{fig7}b) showed a few similarities ($\rho \approx 0.39$, Supplementary Table 5) with the fifth CCA mode (Supplementary Fig. 9e), which indicates that this mode could be either a spurious association or a brain-specific factor that CCA is not able to explicitly identify. This finding indicates the importance of separating the shared factors from the modality-specific ones and the use of more robust inference methods. Furthermore, the relevance of the modality-specific associations would have been more evident if we had included more than two data modalities, where associations within subsets of data modalities could be identified.

Finally, our GFA extension was able to predict a few non-imaging measures from brain connectivity in incomplete data sets. Even though the relative MSE values were modest, the model could predict several measures better than chance. Importantly, the best predicted measures corresponded to the loadings most informative in the shared factors (i.e., the highest absolute loadings), which demonstrates the potential of GFA as a predictive model.

Although the findings from both synthetic and real datasets were robust, there are still a few inherent limitations in our GFA extension. Firstly, the number of initial latent factors $K$ needs to be chosen; however, we have shown in synthetic data that the model can still converge to a good solution even if the number of latent factors is overestimated (Supplementary Fig. 1). Secondly, although the criteria used to select the most relevant factors were validated on synthetic data, these can be further improved, e.g., by including the stability of the factors across multiple initialisations. Thirdly, our GFA extension is computationally demanding to run experiments with incomplete data sets (e.g., the CPU time was approximately 50 hours per initialisation in the HCP experiments). 

Future work should investigate GFA with more data modalities, which could potentially uncover other interesting multivariate associations and improve the predictions of the non-observed data modalities and missing data. Moreover, strategies to improve the interpretability of the factor loadings (e.g., adding additional priors to impose sparsity simultaneously on the group and variable-level) could be implemented. Additionally, automatic inference methods – such as Hamiltonian Monte Carlo or Automatic Differentiation Variational Inference – could be implemented, as these would provide a more flexible framework, permitting new extensions of the model without the need to derive new inference equations. Finally, further extensions of the generative description of GFA could be investigated to improve its predictive accuracy.

\section{Conclusions}
In this study, we have shown that GFA provides an integrative and robust framework that can be used to explore associations among multiple data modalities (in benchmark datasets, such as HCP) and/or predict non-observed data modalities from the observed ones, even if data is missing in one or more data modalities. Due to its Bayesian nature, GFA provides great flexibility to be extended to more complex models to solve more complex tasks, for instance, in neuroscience.

\vspace{12pt}
\textbf{CRediT authorship contribution statement}
\vspace{6pt}

\textbf{Fabio S. Ferreira:} Conceptualization, Methodology, Software, Validation, Formal Analysis, Investigation, Data Curation, Writing - Original Draft, Visualization. \textbf{Agoston Mihalik:} Software, Data Curation, Writing - Review \& Editing, Visualization. \textbf{Rick A. Adams:} Validation, Writing - Review \& Editing. \textbf{John Ashburner}: Conceptualization, Writing - Review \& Editing, Supervision. \textbf{Janaina Mourao-Miranda}: Conceptualization, Remodalities, Writing - Review \& Editing, Supervision, Project administration, Funding acquisition. 

\vspace{12pt}
\textbf{Acknowledgments}
\vspace{6pt}

FSF was supported by \textit{Funda\c{c}\~{a}o para a Ci\^{e}ncia e a Tecnologia} (Ph.D. fellowship No. SFRH/BD/120640/2016). AM, and JM-M were supported by the Wellcome Trust under Grant No. WT102845/Z/13/Z. RAA was supported by a Medical Research Council (MRC) Skills Development Fellowship (Grant No. MR/ S007806/1). The Wellcome Centre for Human Neuroimaging is supported by core funding from the Wellcome Trust (203147/Z/16/Z). Data were provided in part by the Human Connectome Project, WU-Minn Consortium (Principal Investigators: David Van Essen and Kamil Ugurbil; 1U54MH091657) funded by the 16 NIH Institutes and Centers that support the NIH Blueprint for Neuroscience Research; and by the McDonnell Center for Systems Neuroscience at Washington University.

\vspace{12pt}
\textbf{Conflicts of interest}
\vspace{6pt}

The authors do not have any conflicts of interest to disclose.

\begin{appendices}

\section{Variational updates of GFA}
\label{App1}
The variational updates of the model parameters are derived by writing the log of the joint distribution $p(\mathbf{X},\boldsymbol{\theta})$  with respect to all other variational posteriors (Eq. (\ref{eq10})). Considering Eq. (\ref{eq8}), the log of the joint distribution is defined as follows:
\begin{equation} \label{eqa1}
    \ln p(\mathbf{X},\mathbf{Z},\mathbf{W},\boldsymbol{\alpha},\boldsymbol{\tau}) = \ln [p(\mathbf{X}|\mathbf{Z},\mathbf{W},\boldsymbol{\tau})p(\mathbf{Z})p(\mathbf{W}|\boldsymbol{\alpha}) p(\boldsymbol{\alpha})p(\boldsymbol{\tau})] + \text{const},
\end{equation}
where the individual log-densities (considering the priors in Eq. (\ref{eq6}) and Eq. (\ref{eq7})) are given by:
\begin{equation} \label{eqa2} \small
\begin{aligned} 
    \ln p(\mathbf{X}|\mathbf{Z,W},\boldsymbol{\tau}) = & \sum_{m=1}^{M} \bigg[ \frac{N}{2} \sum_{j=1}^{D_{m}} (\ln \tau_{j}^{(m)} - \ln(2\pi)) - \frac{1}{2} \sum_{n=1}^{N} (\mathbf{x}_{n}^{(m)} - \mathbf{W}^{(m)}\mathbf{z}_{n})^{T} \mathbf{T}^{(m)} (\mathbf{x}_{n}^{(m)} - \mathbf{W}^{(m)}\mathbf{z}_{n}) \bigg], \\
    \ln p(\mathbf{Z}) = & -\frac{1}{2} \sum_{n=1}^{N} \mathbf{z}_{n}^{T} \mathbf{z}_{n} - \frac{NK}{2} \ln (2\pi), \\
    \ln p(\mathbf{W}|\boldsymbol{\alpha}) = & \sum_{m=1}^{M} \bigg[ \frac{D_{m}}{2} \sum_{k=1}^{K} \ln \alpha^{(m)}_{k} - \frac{1}{2} \sum_{k=1}^{K} \alpha^{(m)}_{k} \mathbf{w}^{(m) T}_{k} \mathbf{w}^{(m)}_{k} + \frac{D_{m}K}{2} \ln (2\pi) \bigg], \\
    \ln p(\boldsymbol{\alpha}) = & \sum_{m=1}^{M} \sum_{k=1}^{K} \bigg[a_{\boldsymbol{\alpha}^{(m)}} \ln b_{\boldsymbol{\alpha}^{(m)}} - \ln \Gamma(a_{\boldsymbol{\alpha}^{(m)}}) + (a_{\boldsymbol{\alpha}^{(m)}} -1) \ln \alpha^{(m)}_{k} - b_{\boldsymbol{\alpha}^{(m)}} \alpha^{(m)}_{k} \bigg], \\
    \ln p(\boldsymbol{\tau}) = & \sum_{m=1}^{M} \sum_{j=1}^{D_{m}} \bigg[a_{\boldsymbol{\tau}^{(m)}} \ln b_{\boldsymbol{\tau}^{(m)}} - \ln \Gamma(a_{\boldsymbol{\tau}^{(m)}}) + (a_{\boldsymbol{\tau}^{(m)}} -1) \ln \tau^{(m)}_{j} - b_{\boldsymbol{\tau}^{(m)}} \tau^{(m)}_{j} \bigg], \\
\end{aligned}    
\end{equation}
where $\mathbf{T}^{(m)} = \text{diag}(\boldsymbol{\tau}^{(m)})$, $\mathbf{z}_{n}$ is the $n$-th column of $\mathbf{Z}$, $\mathbf{x}_{n}^{(m)}$ is the $n$-th column of $\mathbf{X}^{(m)}$ and $\mathbf{w}_{k}^{(m)}$ is a column vector representing the $k$-th column of $\mathbf{W}^{(m)}$. 

\subsection{$q(\mathbf{Z})$ distribution}
The optimal log-density for $q(\mathbf{Z})$, given the other variational distributions is calculated using Eq. (\ref{eq10}):
\begin{equation} \label{eqa3} \small
\begin{aligned}
    \ln q(\mathbf{Z}) =  & \ \mathbb{E}_{q(\mathbf{W}), q(\boldsymbol{\tau})} [\ln p(\mathbf{X}|\mathbf{Z,W,\tau}) + \ln p(\mathbf{Z})], \\
                  =  & \sum_{n=1}^{N} \bigg[ -\frac{1}{2} \sum_{m=1}^{M} \langle (\mathbf{x}_{n} - \mathbf{W}^{(m)}\mathbf{z}_{n})^{T} \mathbf{T}^{(m)} (\mathbf{x}_{n} - \mathbf{W}^{(m)}\mathbf{z}_{n}) \rangle -\frac{1}{2} \mathbf{z}_{n}^{T} \mathbf{z}_{n} \bigg], \\
                  =  & \sum_{n=1}^{N} \bigg[ \mathbf{z}_{n}^{T} \sum_{m=1}^{M} \sum_{j \in O_{n}^{(m)}} \langle \tau_{j}^{(m)} \rangle \langle \mathbf{w}_{j}^{(m)^{T}} \rangle x_{j,n} -\frac{1}{2} \mathbf{z}_{n}^{T} \bigg( \sum_{m=1}^{M} \sum_{j \in O_{n}^{(m)}} \langle \tau_{j}^{(m)} \rangle \langle \mathbf{w}_{j}^{(m)^{T}} \mathbf{w}_{j}^{(m)} \rangle \bigg) \mathbf{z}_{n} -\frac{1}{2} \mathbf{z}_{n}^{T} \mathbf{z}_{n} \bigg], \\
                  =  & \sum_{n=1}^{N} \bigg[ \mathbf{z}_{n}^{T} \sum_{m=1}^{M} \sum_{j \in O_{n}^{(m)}} \langle \tau_{j}^{(m)} \rangle \langle \mathbf{w}_{j}^{(m)^{T}} \rangle x_{j,n} -\frac{1}{2} \mathbf{z}_{n}^{T} \bigg( \mathbf{I}_{K} + \sum_{m=1}^{M} \sum_{j \in O_{n}^{(m)}} \langle \tau_{j}^{(m)} \rangle \langle \mathbf{w}_{j}^{(m)^{T}} \mathbf{w}_{j}^{(m)} \rangle \bigg) \mathbf{z}_{n} \bigg],
\end{aligned}    
\end{equation}
where $\langle \cdot \rangle = \mathbb{E}_{q(\mathbf{W}), q(\boldsymbol{\tau})} [\cdot]$ represents expectations, $\mathbf{w}^{(m)}_{j}$ denotes the $j$-th row of $\mathbf{W}^{(m)}$, $\langle \tau_{j}^{(m)} \rangle = \tfrac{\Tilde{a}^{(j)}_{\boldsymbol{\tau}^{(m)}}}{\Tilde{b}^{(j)}_{\boldsymbol{\tau}^{(m)}}}$ ($\Tilde{a}^{(j)}_{\boldsymbol{\tau}^{(m)}}$ and $\Tilde{b}^{(j)}_{\boldsymbol{\tau}^{(m)}}$ are the variational parameters obtained for $q(\boldsymbol{\tau}^{(m)})$ in Eq. (\ref{eqa17})) and $\langle \mathbf{w}_{j}^{(m)^{T}} \mathbf{w}_{j}^{(m)} \rangle = \mathbf{\Sigma}_{\mathbf{w}^{(m)}_{j}} + \boldsymbol{\mu}_{\mathbf{w}_{j}^{(m)}}^{T} \boldsymbol{\mu}_{\mathbf{w}_{j}^{(m)}}$ ($\mathbf{\Sigma}_{\mathbf{w}_{j}^{(m)}}$ and $\boldsymbol{\mu}_{\mathbf{w}_{j}^{(m)}}$ are the variational parameters obtained for $q(\mathbf{W}^{(m)})$ in Eq. (\ref{eqa11})). $O_{n}^{(m)}$ is the set of indices in the $n$-th column of $\mathbf{X}^{(m)}$ ($x_{:,n}^{(m)}$) that are not missing. In Eq. (\ref{eqa3}), we have omitted constant terms that do not depend on $\mathbf{Z}$. Taking the exponential of the log density, the optimal $q(\mathbf{Z})$ is a multivariate normal distribution:
\begin{equation} \label{eqa4}
    q(\mathbf{Z}) = \prod_{n=1}^{N} q(\mathbf{z}_{n}) = \prod_{n=1}^{N} \mathcal{N} (\mathbf{z}_{n}| \boldsymbol{\mu}_{ \mathbf{z}_{n}}, \mathbf{\Sigma}_{\mathbf{z}_{n}}).
\end{equation}

The updates equations for $q(\mathbf{Z})$ are:
\begin{equation} \label{eqa5}
\begin{gathered}
    \mathbf{\Sigma}_{\mathbf{z}_{n}} = \bigg[ \mathbf{I}_{K} + \sum_{m=1}^{M} \ \sum_{j \in O_{n}^{(m)}} \langle \tau_{j}^{(m)} \rangle \langle \mathbf{w}_{j}^{(m)^{T}} \mathbf{w}_{j}^{(m)} \rangle\bigg]^{-1}, \\
    \boldsymbol{\mu}_{\mathbf{z}_{n}} = \mathbf{\Sigma}_{\mathbf{z}_{n}} \sum_{m=1}^{M} \ \sum_{j \in O_{n}^{(m)}} \langle \tau_{j}^{(m)} \rangle \langle \mathbf{w}_{j}^{(m)^{T}} \rangle x_{j,n}.
\end{gathered}    
\end{equation}

\subsection{$q(\mathbf{W}^{(m)})$ distribution}
The optimal log-density for $q(\mathbf{W}^{(m)})$, given the other variational distributions is obtained by calculating:
\begin{equation} \label{eqa6}
\begin{aligned}
    \ln q(\mathbf{W}^{(m)}) =  & \ \mathbb{E}_{q(\mathbf{Z}),q(\boldsymbol{\alpha}^{(m)}),q(\boldsymbol{\tau}^{(m)})} [\ln p(\mathbf{X}^{(m)}|\mathbf{Z},\mathbf{W}^{(m)},\boldsymbol{\tau}^{(m)}) + \ln p(\mathbf{W}^{(m)}|\boldsymbol{\alpha}^{(m)})], \\
                  =  & -\frac{1}{2} \sum_{n=1}^{N} \langle (\mathbf{x}_{n} - \mathbf{W}^{(m)}\mathbf{z}_{n})^{T} \mathbf{T}^{(m)} (\mathbf{x}_{n} - \mathbf{W}^{(m)}\mathbf{z}_{n}) \rangle -\frac{1}{2} \sum_{k=1}^{K} \langle \alpha_{k}^{(m)} \mathbf{w}_{k}^{(m) T} \mathbf{w}_{k}^{(m)} \rangle, \\
\end{aligned}    
\end{equation}
where $\langle \cdot \rangle = \mathbb{E}_{q(\mathbf{Z}),q(\boldsymbol{\alpha}^{(m)}),q(\boldsymbol{\tau}^{(m)})} [\cdot]$. The constant term was omitted. The first term of Eq. (\ref{eqa6}) can be expanded as follows:

\begin{equation} \label{eqa7}
\begin{aligned}
    & -\frac{1}{2} \sum_{n=1}^{N} \langle (\mathbf{x}_{n} - \mathbf{W}^{(m)}\mathbf{z}_{n})^{T} \mathbf{T}^{(m)} (\mathbf{x}_{n} - \mathbf{W}^{(m)}\mathbf{z}_{n}) \rangle,\\
    & = \sum_{j=1}^{D_{m}} \langle \tau_{j}^{(m)} \rangle \bigg( \sum_{n \in O_{j}^{(m)}} x_{j,n}^{(m)} \langle \mathbf{z}_{n}^{T} \rangle \bigg) \mathbf{w}_{j}^{(m)^{T}} + \sum_{j=1}^{D_{m}} -\frac{1}{2} \mathbf{w}_{j}^{(m)} \bigg( \langle \tau_{j}^{(m)} \rangle \sum_{n \in O_{j}^{(m)}} \langle \mathbf{z}_{n} \mathbf{z}_{n}^{T} \rangle \bigg) \mathbf{w}_{j}^{(m)^{T}},  
\end{aligned}    
\end{equation}
where $\langle \mathbf{z}_{n} \mathbf{z}_{n}^{T} \rangle = \mathbf{\Sigma}_{\mathbf{z}_{n}} + \boldsymbol{\mu}_{\mathbf{z}_{n}} \boldsymbol{\mu}_{\mathbf{z}_{n}}^{T}$ ($\mathbf{\Sigma}_{\mathbf{z}_{n}}$ and $\boldsymbol{\mu}_{\mathbf{z}_{n}}$ are the variational parameters of $q(\mathbf{Z})$ in Equation \ref{eqa5}) and $O_{j}^{(m)}$ is the set of indices in the $j$-th row of $\mathbf{X}^{(m)}$ ($x_{j,:}^{(m)}$) that are not missing. The second term of Eq. (\ref{eqa6}) is given by: 
\begin{equation} \label{eqa8}
    -\frac{1}{2} \sum_{k=1}^{K} \langle \alpha_{k}^{(m)} \mathbf{w}_{k}^{(m) T} \mathbf{w}_{k}^{(m)} \rangle = -\frac{1}{2} \sum_{j=1}^{D_{m}} \mathbf{w}_{j}^{(m)} \langle \mathbf{A}_{\alpha}^{(m)} \rangle \mathbf{w}_{j}^{(m)^{T}},
\end{equation}
where $\langle \mathbf{A}_{\alpha}^{(m)} \rangle = \text{diag} (\langle \boldsymbol{\alpha}^{(m)} \rangle)$ and $ \langle \boldsymbol{\alpha}^{(m)} \rangle = \tfrac{\Tilde{a}_{\boldsymbol{\alpha}^{(m)}}}{\Tilde{\mathbf{b}}_{\boldsymbol{\alpha}^{(m)}}}$ ($\Tilde{a}_{\boldsymbol{\alpha}^{(m)}}$ and $\Tilde{\mathbf{b}}_{\boldsymbol{\alpha}^{(m)}}$ are the variational parameters of $q(\boldsymbol{\alpha}^{(m)})$ in Eq. (\ref{eqa14})). Putting both terms together we get:
\begin{equation} \label{eqa9}
\begin{aligned}
    \ln q(\mathbf{W}^{(m)}) =  & \sum_{j=1}^{D_{m}} \bigg[ \langle \tau_{j}^{(m)} \rangle \bigg( \sum_{n \in O_{j}^{(m)}} x_{j,n}^{(m)} \langle \mathbf{z}_{n}^{T} \rangle \bigg) \mathbf{w}_{j}^{(m)^{T}}, \\
    & -\frac{1}{2} \mathbf{w}_{j}^{(m)} \bigg( \langle \mathbf{A}_{\alpha}^{(m)} \rangle + \langle \tau_{j}^{(m)} \rangle \sum_{n \in O_{j}^{(m)}} \langle \mathbf{z}_{n} \mathbf{z}_{n}^{T} \rangle \bigg) \mathbf{w}_{j}^{(m)^{T}} \bigg].
\end{aligned}    
\end{equation}

Taking the exponential of the log density, the optimal $q(\mathbf{W}^{(m)})$ is a multivariate normal distribution:
\begin{equation} \label{eqa10}
    q(\mathbf{W}^{(m)}) = \prod_{j=1}^{D_{m}} q(\mathbf{w}_{j}^{(m)}) = \prod_{j=1}^{D_{m}} \mathcal{N} (\mathbf{w}_{j}^{(m)}| \boldsymbol{\mu}_{\mathbf{w}^{(m)}_{j}}, \mathbf{\Sigma}_{\mathbf{w}^{(m)}_{j}}).
\end{equation}

Then the updates equations for $q(\mathbf{W}^{(m)})$ are:
\begin{equation} \label{eqa11}
\begin{gathered}
    \mathbf{\Sigma}_{\mathbf{w}^{(m)}_{j}} = \bigg[ \langle \mathbf{A}_{\alpha}^{(m)} \rangle + \langle \tau_{j}^{(m)} \rangle \sum_{n \in O_{j}^{(m)}} \langle \mathbf{z}_{n} \mathbf{z}_{n}^{T} \rangle\bigg]^{-1}, \\
    \boldsymbol{\mu}_{\mathbf{w}^{(m)}_{j}} = \langle \tau_{j}^{(m)} \rangle \sum_{n \in O_{j}^{(m)}} \bigg( x_{j,n} \langle \mathbf{z}_{n}^{T} \rangle \bigg) \mathbf{\Sigma}_{\mathbf{w}^{(m)}_{j}}.
\end{gathered}    
\end{equation}

\subsection{\texorpdfstring{$q(\boldsymbol{\alpha}^{(m)})$}{TEXT} distribution}
The optimal log-density for $q(\boldsymbol{\alpha}^{(m)})$, given the other variational distributions is obtained by calculating:
\begin{equation} \label{eqa12}
\begin{aligned}
    \ln q(\boldsymbol{\alpha}^{(m)}) =  & \ \mathbb{E}_{q(\mathbf{W}^{(m)})} [\ln p(\mathbf{W^{(m)}|\alpha^{(m)}}) + \ln p(\boldsymbol{\alpha}^{(m)})], \\
                  =  & \sum_{k=1}^{K} \bigg[ \frac{D_{m}}{2} \ln \alpha^{(m)}_{k} - \frac{1}{2} \alpha^{(m)}_{k} \langle  \mathbf{w}^{(m)^{T}}_{k} \mathbf{w}^{(m)}_{k} \rangle + (a_{\boldsymbol{\alpha}^{(m)}} -1) \ln \alpha^{(m)}_{k} - b_{\boldsymbol{\alpha}^{(m)}} \alpha^{(m)}_{k} \bigg], \\
                  =  & \sum_{k=1}^{K} \bigg( \frac{D_{m}}{2} + a_{\boldsymbol{\alpha}^{(m)}} -1 \bigg) \ln \alpha^{(m)}_{k} - \sum_{k=1}^{K} \bigg( b_{\boldsymbol{\alpha}^{(m)}} + \frac{1}{2}  \langle  \mathbf{w}^{(m)^{T}}_{k} \mathbf{w}^{(m)}_{k} \rangle \bigg) \alpha^{(m)}_{k},
\end{aligned}    
\end{equation}
where $\langle \cdot \rangle = \mathbb{E}_{q(\mathbf{W}^{(m)})} [\cdot]$. We have omitted constant terms that do not depend on $\boldsymbol{\alpha}$. Taking the exponential of the log density, the optimal $q(\boldsymbol{\alpha}^{(m)})$ is a Gamma distribution:
\begin{equation} \label{eqa13}
    q(\boldsymbol{\alpha}^{(m)}) = \prod_{k=1}^{K} q(\boldsymbol{\alpha}_{k}^{(m)}) = \prod_{k=1}^{K} \Gamma(\alpha^{(m)}_{k}| \Tilde{a}_{\boldsymbol{\alpha}^{(m)}}, \Tilde{b}^{(k)}_{\boldsymbol{\alpha}^{(m)}}).
\end{equation}

And the update equations for $q(\boldsymbol{\alpha}^{(m)})$ are:
\begin{equation} \label{eqa14}
\begin{gathered}
    \Tilde{a}_{\boldsymbol{\alpha}^{(m)}} = a_{\boldsymbol{\alpha}^{(m)}} + \frac{1}{2}D_{m}, \\
    \Tilde{b}^{(k)}_{\boldsymbol{\alpha}^{(m)}} = b_{\boldsymbol{\alpha}^{(m)}} + \frac{1}{2}  \langle  \mathbf{w}^{(m)^{T}}_{k} \mathbf{w}^{(m)}_{k} \rangle,
\end{gathered}    
\end{equation}
where $\langle \mathbf{w}^{(m) T}_{k} \mathbf{w}^{(m)}_{k} \rangle = \bigg( \sum_{j=1}^{D_{m}} \boldsymbol{\mu}_{\mathbf{w}_{j}^{(m)}}^{T} \boldsymbol{\mu}_{\mathbf{w}_{j}^{(m)}} + \mathbf{\Sigma}_{\mathbf{w}_{j}^{(m)}} \bigg)_{(k,k)}$.

\subsection{\texorpdfstring{$q(\boldsymbol{\tau}^{(m)})$}{TEXT} distribution}
The optimal log-density for $q(\boldsymbol{\tau}^{(m)})$, given the other variational distributions is obtained in the following way:
\begin{equation} \label{eqa15}
\begin{aligned}
    \ln q(\boldsymbol{\tau}^{(m)}) =  & \ \mathbb{E}_{q(\mathbf{Z}),q(\mathbf{W}^{(m)})} [\ln p(\mathbf{X}^{(m)}|\mathbf{Z},\mathbf{W}^{(m)},\boldsymbol{\tau}^{(m)}) + \ln p(\boldsymbol{\tau}^{(m)})]\\
                  =  & -\frac{1}{2} \sum_{j=1}^{D_{m}} \bigg[ \tau_{j}^{(m)} \sum_{n \in O_{j}^{(m)}} \bigg( x_{j,n}^{(m) 2} -2 x_{j,n}^{(m)} \langle \mathbf{w}_{j}^{(m)} \rangle \langle \mathbf{z}_{n} \rangle + \text{Tr}[\langle \mathbf{w}_{j}^{(m)^{T}} \mathbf{w}_{j}^{(m)} \rangle \langle \mathbf{z}_{n} \mathbf{z}_{n}^{T} \rangle]  \bigg)\\
                  & + \frac{N_{j}^{(m)}}{2} \ln \tau_{j}^{(m)} + (a_{\boldsymbol{\tau}^{(m)}} -1) \ln \tau^{(m)}_{j} - b_{\boldsymbol{\tau}^{(m)}} \tau^{(m)}_{j} \bigg] \\
                  =  & \sum_{j=1}^{D_{m}} \bigg(a_{\boldsymbol{\tau}^{(m)}} + \frac{N_{j}^{(m)}}{2} -1\bigg) \ln \tau^{(m)}_{j} - \sum_{j=1}^{D_{m}} \bigg( b_{\boldsymbol{\tau}^{(m)}} + \frac{1}{2} \sum_{n \in O_{j}^{(m)}} x_{j,n}^{(m) 2} - 2 x_{j,n}^{(m)} \langle \mathbf{w}_{j}^{(m)} \rangle \langle \mathbf{z}_{n} \rangle  \\
                  & + \text{Tr}[\langle \mathbf{w}_{j}^{(m)^{T}} \mathbf{w}_{j}^{(m)} \rangle \langle \mathbf{z}_{n} \mathbf{z}_{n}^{T} \rangle] \bigg) \tau_{j}^{(m)},   
\end{aligned}    
\end{equation}
where $N_{j}^{(m)}$ is the number of non-missing observations in the $j$-th row of the corresponding group $\mathbf{X}^{(m)}$ and $\langle \cdot \rangle = \mathbb{E}_{q(\mathbf{Z}),q(\mathbf{W}^{(m)})} [\cdot]$. We have omitted constant terms that do not depend on $\tau$. Taking the exponential of the log density, the optimal $q(\boldsymbol{\tau}^{(m)})$ is a Gamma distribution:
\begin{equation}
    q(\boldsymbol{\tau}^{(m)}) = \prod_{j=1}^{D_{m}} q(\boldsymbol{\tau}_{j}^{(m)}) = \prod_{j=1}^{D_{m}} \Gamma(\tau^{(m)}_{j}| \Tilde{a}^{(j)}_{\boldsymbol{\tau}^{(m)}}, \Tilde{b}^{(j)}_{\boldsymbol{\tau}^{(m)}}),
\end{equation}
where the variational parameters are calculated by:
\begin{equation}
\label{eqa17}
\begin{gathered}
    \Tilde{a}^{(j)}_{\boldsymbol{\tau}^{(m)}} = a_{\boldsymbol{\tau}^{(m)}} + \frac{1}{2}N_{j}^{(m)}, \\
    \Tilde{b}^{(j)}_{\boldsymbol{\tau}^{(m)}} = b_{\boldsymbol{\tau}^{(m)}} + \frac{1}{2} \sum_{n \in O_{j}^{(m)}} x_{j,n}^{(m) 2} - 2 x_{j,n}^{(m)} \langle \mathbf{w}_{j}^{(m)} \rangle \langle \mathbf{z}_{n} \rangle  + \text{Tr}[\langle \mathbf{w}_{j}^{(m)^{T}} \mathbf{w}_{j}^{(m)} \rangle \langle \mathbf{z}_{n} \mathbf{z}_{n}^{T} \rangle].
\end{gathered}
\end{equation}

Finally, to solve the rotation and scaling ambiguity known to be present in factor analysis models, we used a similar approach previously proposed by Virtanen and colleagues \citep{Virtanen2011, Virtanen2012, Klami2013}, which consists of maximising the variational lower bound with respect to a linear transformation $\mathbf{R}$ of the latent space, after each round of variational EM updates. This also improves convergence and speeds up the learning.   

\section{Lower bound for GFA}
\label{App2}
Considering Eq. (\ref{eq9}), the lower bound of $\ln p(\mathbf{X})$ is given by:
\begin{equation}
\begin{aligned}
     \mathcal{L}(q) =  & \ \mathbb{E} [\ln p(\mathbf{X, Z,W,\alpha,\tau})] - \mathbb{E} [\ln q(\mathbf{Z,W,\alpha,\tau})]\\
                  =  & \ \mathbb{E} [\ln p(\mathbf{X}|\mathbf{Z,W,\tau})] + \mathbb{E} [\ln p(\mathbf{Z})] + \mathbb{E} [\ln p(\mathbf{W}|\boldsymbol{\alpha})] + \mathbb{E} [\ln p(\boldsymbol{\alpha})] + \mathbb{E} [\ln p(\boldsymbol{\tau})]\\
                  - & \ \mathbb{E} [\ln q(\mathbf{Z})] + \mathbb{E} [\ln q(\mathbf{W})] + \mathbb{E} [\ln q(\boldsymbol{\alpha})] + \mathbb{E} [\ln q(\boldsymbol{\tau})],
\end{aligned}    
\end{equation}
where the expectations of the $\ln p(\cdot)$ terms are given by (see Eq. (\ref{eqa2})):
\begin{equation}
    \mathbb{E}_{q(\boldsymbol{\theta})} [\ln p(\mathbf{X}|\mathbf{Z,W,\tau})] = \sum_{m=1}^{M} \bigg[ \sum_{j=1}^{D_{m}} \bigg( \frac{N_{j}^{(m)}}{2} (\langle \ln \tau_{j}^{(m)} \rangle - \ln (2\pi)) - \langle \tau_{j}^{(m)} \rangle (\Tilde{b}^{(j)}_{\boldsymbol{\tau}^{(m)}} - b_{\boldsymbol{\tau}^{(m)}}) \bigg) \bigg],
\end{equation}
\begin{equation}
    \mathbb{E} [\ln p(\mathbf{Z})] = -\frac{1}{2} \sum_{n=1}^{N} \text{Tr}[\langle \mathbf{z}_{n} \mathbf{z}_{n}^{T} \rangle] - \frac{NK}{2} \ln (2\pi),
\end{equation}
\begin{equation}
    \mathbb{E}_{q(\boldsymbol{\alpha})} [\ln p(\mathbf{W}|\boldsymbol{\alpha})] = \sum_{m=1}^{M} \bigg[ \frac{D_{m}}{2} \sum_{k=1}^{K} \langle \ln \alpha_{k}^{(m)} \rangle - \sum_{k=1}^{K} \text{Tr}[\langle \alpha_{k}^{(m)}\rangle \langle \mathbf{w}_{k}^{(m)^{T}} \mathbf{w}_{k}^{(m)} \rangle] + \frac{D_{m}K}{2} \ln (2\pi) \bigg],
\end{equation}
\begin{equation}
    \mathbb{E} [\ln p(\boldsymbol{\alpha})] = \sum_{m=1}^{M} \sum_{k=1}^{K} \bigg[a_{\boldsymbol{\alpha}^{(m)}} \ln b_{\boldsymbol{\alpha}^{(m)}} - \ln \Gamma(a_{\boldsymbol{\alpha}^{(m)}}) + (a_{\boldsymbol{\alpha}^{(m)}} -1) \langle \ln \alpha^{(m)}_{k} \rangle - b_{\boldsymbol{\alpha}^{(m)}} \langle \alpha^{(m)}_{k} \rangle \bigg],
\end{equation}
\begin{equation}
    \mathbb{E} [\ln p(\boldsymbol{\tau})] = \sum_{m=1}^{M} \sum_{j=1}^{D_{m}} \bigg[a_{\boldsymbol{\tau}^{(m)}} \ln b_{\boldsymbol{\tau}^{(m)}} - \ln \Gamma(a_{\boldsymbol{\tau}^{(m)}}) + (a_{\boldsymbol{\tau}^{(m)}} -1) \langle \ln \tau^{(m)}_{j} \rangle - b_{\boldsymbol{\tau}^{(m)}} \langle \tau^{(m)}_{j} \rangle \bigg],
\end{equation}
where $q(\boldsymbol{\theta}) = q(\mathbf{Z})q(\mathbf{W})q(\boldsymbol{\tau})$, $\langle \ln \tau_{j}^{(m)} \rangle = \psi(\Tilde{a}^{(j)}_{\boldsymbol{\tau}^{(m)}}) - \ln \Tilde{b}^{(j)}_{\boldsymbol{\tau}^{(m)}}$, $\langle \ln \alpha_{k}^{(m)} \rangle = \psi(\Tilde{a}_{\boldsymbol{\alpha}^{(m)}}) - \ln \Tilde{b}^{(k)}_{\boldsymbol{\alpha}^{(m)}}$ and $\psi(\cdot)$ is a digamma function. $\langle \tau_{j}^{(m)} \rangle$, $\langle \mathbf{z}_{n} \mathbf{z}_{n}^{T} \rangle$, $\langle \alpha_{k}^{(m)} \rangle$ and $\langle \mathbf{w}^{(m) T}_{k} \mathbf{w}^{(m)}_{k} \rangle$ are calculated as in Eq. (\ref{eqa3}), Eq. (\ref{eqa7}), Eq. (\ref{eqa8}) and Eq. (\ref{eqa14}), respectively.

The terms involving expectations of the logs of the $q(\cdot)$ distributions simply represent the negative entropies of those distributions \citep{Bishop2006}:
\begin{equation}
    \mathbb{E} [\ln q(\mathbf{Z})] = - \frac{1}{2} \bigg[ \sum_{n=1}^{N} \ln |\Sigma_{\mathbf{z}_{n}}| + K(1 + \ln (2\pi)) \bigg], 
\end{equation}
\begin{equation}
    \mathbb{E} [\ln q(\mathbf{W})] = \sum_{m=1}^{M} - \frac{1}{2} \bigg[ \sum_{j=1}^{D_{m}} \ln |\Sigma_{\mathbf{w}_{j}^{(m)}}| + K(1 + \ln (2\pi)) \bigg],
\end{equation}
\begin{equation}
    \mathbb{E} [\ln q(\boldsymbol{\alpha})] = \sum_{m=1}^{M} \sum_{k=1}^{K} \bigg[ \Tilde{a}_{\boldsymbol{\alpha}^{(m)}} \ln \Tilde{b}^{(k)}_{\boldsymbol{\alpha}^{(m)}} - \ln \Gamma(\Tilde{a}_{\boldsymbol{\alpha}^{(m)}}) + (\Tilde{a}_{\boldsymbol{\alpha}^{(m)}} -1) \langle \ln \alpha^{(m)}_{k} \rangle - \Tilde{b}^{(k)}_{\boldsymbol{\alpha}^{(m)}} \langle \alpha^{(m)}_{k}  \rangle \bigg], 
\end{equation}
\begin{equation} 
    \mathbb{E} [\ln q(\boldsymbol{\tau})] = \sum_{m=1}^{M} \sum_{j=1}^{D_{m}} \bigg[ \Tilde{a}^{(j)}_{\boldsymbol{\tau}^{(m)}} \ln \Tilde{b}^{(j)}_{\boldsymbol{\tau}^{(m)}} - \ln \Gamma(\Tilde{a}^{(j)}_{\boldsymbol{\tau}^{(m)}}) + (\Tilde{a}^{(j)}_{\boldsymbol{\tau}^{(m)}} -1) \langle \ln \tau^{(m)}_{j} \rangle - \Tilde{b}^{(j)}_{\boldsymbol{\tau}^{(m)}} \langle \tau^{(m)}_{j}  \rangle \bigg].
\end{equation}
\end{appendices}

\bibliographystyle{elsarticle-harv}
\bibliography{main}

\cleardoublepage

\textbf{\Large Supplementary Material: A hierarchical Bayesian model to find brain-behaviour associations in incomplete data sets}

\vspace{6pt}
\textbf{Materials and Methods}
\vspace{6pt}

\textit{Additional GFA experiments on synthetic data}
\vspace{6pt}

We ran GFA experiments on the following selections of synthetic data:
\begin{enumerate}
    \item Complete data (all models were  initialised with $K=30$):
    \begin{enumerate}
        \item low dimensional data ($D_{1} = 50$ and $D_{2} = 30$) was generated using the same parameters described in Section \ref{methods_expsynt}.
        \item high dimensional data was generated ($D_{1} = 20000$ and $D_{2} = 200$) using the same parameters described in Section \ref{methods_expsynt}. 
    \end{enumerate}
    \item Incomplete data (all models were  initialised with $K=15$):
    \begin{enumerate}
        \item the elements of $\mathbf{X}^{(2)}$ deviating more than $1\sigma$ (i.e., standard deviation) from the mean (i.e, $x_{dn} > \mu + \sigma$ and $x_{dn} < \mu - \sigma$) were removed from the synthetic data generated in the supplementary experiment 1a, which led to approximately $30\%$ of missing values in $\mathbf{X}^{(2)}$.
        \item $10\%$ of the rows of $\mathbf{X}^{(1)}$ and $20\%$ of the elements of $\mathbf{X}^{(2)}$ were randomly removed from the low dimensional data generated in the supplementary experiment 1a.
        \item $10\%$ of the rows of $\mathbf{X}^{(1)}$ and $20\%$ of the elements of $\mathbf{X}^{(2)}$ were randomly removed from the high dimensional data generated in the supplementary experiment 1b. 
    \end{enumerate}
\end{enumerate}

\textit{CCA experiments on synthetic data}
\vspace{6pt}

In order to assess the CCA performance in complete and incomplete data sets, we generated data using the parameters described in Section \ref{methods_expsynt} and ran experiments on the following selections of the data:
\begin{itemize}
    \item Complete data
    \item Incomplete data
    \begin{itemize}
        \item $20\%$ of the elements of $\mathbf{X}^{(1)}$ and $40\%$ of the elements of $\mathbf{X}^{(2)}$ were randomly removed.
        \item the elements of $\mathbf{X}^{(1)}$ and $\mathbf{X}^{(2)}$ deviating more than $1\sigma$ from the mean were removed, which led to approximately $30\%$ of missing values in each data modality.
    \end{itemize}
\end{itemize}

The missing values were imputed using the median. The statistical significance of the CCA modes was estimated by permutation inference, in which the rows of $\mathbf{X}^{(2)}$ were permuted 1000 times and CCA was run after each permutation. For each CCA mode, we compute a p-value to assess whether the ``true" canonical correlation (i.e., the canonical correlation of the respective CCA mode obtained without permuting the data) was larger than the null distribution of permuted canonical correlations of the first CCA mode (equivalent to a maximum statistics approach). To obtain an equivalent representation of a single latent variable for CCA (comparable to a latent factor in GFA), the canonical scores $\mathbf{U}^{T}\mathbf{X}^{(1)}$ and $\mathbf{V}^{T}\mathbf{X}^{(2)}$, where $\mathbf{U} \in \mathbb{R}^{D_{1} \times K}$ and $\mathbf{V} \in\mathbb{R}^{D_{2} \times K}$, were averaged. 

These experiments were also run using our GFA extension without imputing the missing values. The performance of the models were assessed by visually comparing the inferred latent factors. The incomplete data experiments were different from those described in Section \ref{methods_expsynt} because we wanted to show the potential of GFA to handle missing data when more than one modality had missing values. Moreover, it would be of little interest to run CCA with missing rows because, in practice, one would not impute the missing values but rather remove the rows in both data modalities.

\vspace{6pt}
\textit{CCA experiments on the HCP data}
\vspace{6pt}

To compare our GFA results with CCA, we applied a CCA analysis similar to the one proposed by \citet{Smith2015} to the HCP data used in the GFA experiments. In summary, we reduced the dimensionality of both data modalities using Principal Component Analysis (fixing the number of principal components in each data modality to 100) and applied CCA to these reduced data matrices. The statistical significance of the CCA modes was estimated by permutation inference, in which the subjects of the non-imaging matrix were permuted 10,000 times respecting the family structure of the data \citep{Winkler2015}. The permutation approach was identical to that described in the \textit{CCA experiments on synthetic data} section. For more details of the analysis, see \citet{Smith2015}.

\vspace{6pt}
\textit{Surface plots}
\vspace{6pt}

\sloppy The surface plots illustrate maps of brain connection strength increases/decreases, which were obtained by weighting each node’s parcel map with the GFA/CCA edge-strengths (the loadings were multiplied by the sign of the population mean correlation) summed across the edges connected to the node. We used the node's parcel maps provided as a cifti file (named \textit{melodic\_IC\_ftb.dlabel.nii}) in the group ICA folder (named \textit{groupICA\_3T\_HCP1200\_MSMAll\_d200.ica}). In this file, one can find the number of the ICA component that each vertex is most likely to belong to.

\vspace{6pt}
\textbf{Results}
\vspace{6pt}

\textit{Additional GFA experiments on synthetic data}
\vspace{6pt}

The model parameters were correctly inferred using low (Supplementary Fig. 1a) ($\hat{\tau}^{(1)} \approx 5.10$ and $\hat{\tau}^{(2)}\approx9.98$) and high (Supplementary Fig. 1b) ($\hat{\tau}^{(1)} \approx 5.01$ and $\hat{\tau}^{(2)} \approx 9.97$) dimensional synthetic data, when the model was initialised with $K=30$. The most relevant shared and modality-specific factors were correctly estimated in both experiments.

In supplementary experiment 2a, taking into account the difficulty of the task our GFA approach recovered the model parameters fairly well ($\hat{\tau}^{(1)} \approx 5.04$ and $\hat{\tau}^{(2)} \approx 11.72$), whereas the median imputation approach failed to estimate the noise parameter of the second modality ($\hat{\tau}^{(1)} \approx 5.03$ and $\hat{\tau}^{(2)} \approx 6.95$) and the third factor (i.e. the factor specific to $\mathbf{X}^{(2)}$) was erroneously identified (Supplementary Fig. 2a). Furthermore, our approach performed better in the multi-output prediction task (Supplementary Table 1). The proposed GFA extension predicted missing data accurately ($\rho = 0.929 \pm 0.021$). 

In the supplementary experiments 2b (Supplementary Fig. 2b) and 2c (Supplementary Fig. 2c), our approach inferred the model parameters correctly in low ($\hat{\tau}^{(1)} \approx 5.01$ and $\hat{\tau}^{(2)} \approx 10.15$) and high dimensional ($\hat{\tau}^{(1)} \approx 5.03$ and $\hat{\tau}^{(2)} \approx 9.97$) data sets, respectively. The median imputation approach failed to estimate the model parameters in both experiments ($\hat{\tau}^{(1)} \approx 6.23$ and $\hat{\tau}^{(2)} \approx 6.39$ in supplementary experiment 2b (Supplementary Fig. 2b); $\hat{\tau}^{(1)} \approx 6.33$ and $\hat{\tau}^{(2)} \approx 3.95$ in supplementary experiment 2c (Supplementary Fig. 2c)). The performance of both approaches in the multi-output prediction task was similar and below chance level (Supplementary Table 1). Our GFA extension predicted reasonably well the missing observations in both modalities (supplementary experiment 2b: $\rho = 0.675 \pm 0.031$ and $\rho = 0.779 \pm 0.022$ for the missing values in $\mathbf{X}^{(1)}$ and $\mathbf{X}^{(2)}$, respectively; supplementary experiment 2c: $\rho = 0.627 \pm 0.012$ and $\rho = 0.859 \pm 0.003$ for the missing values in $\mathbf{X}^{(1)}$ and $\mathbf{X}^{(2)}$, respectively).

\setcounter{figure}{0}
\setcounter{table}{0}

\begin{figure}[H]
\centering
\includegraphics[width=0.70\linewidth]{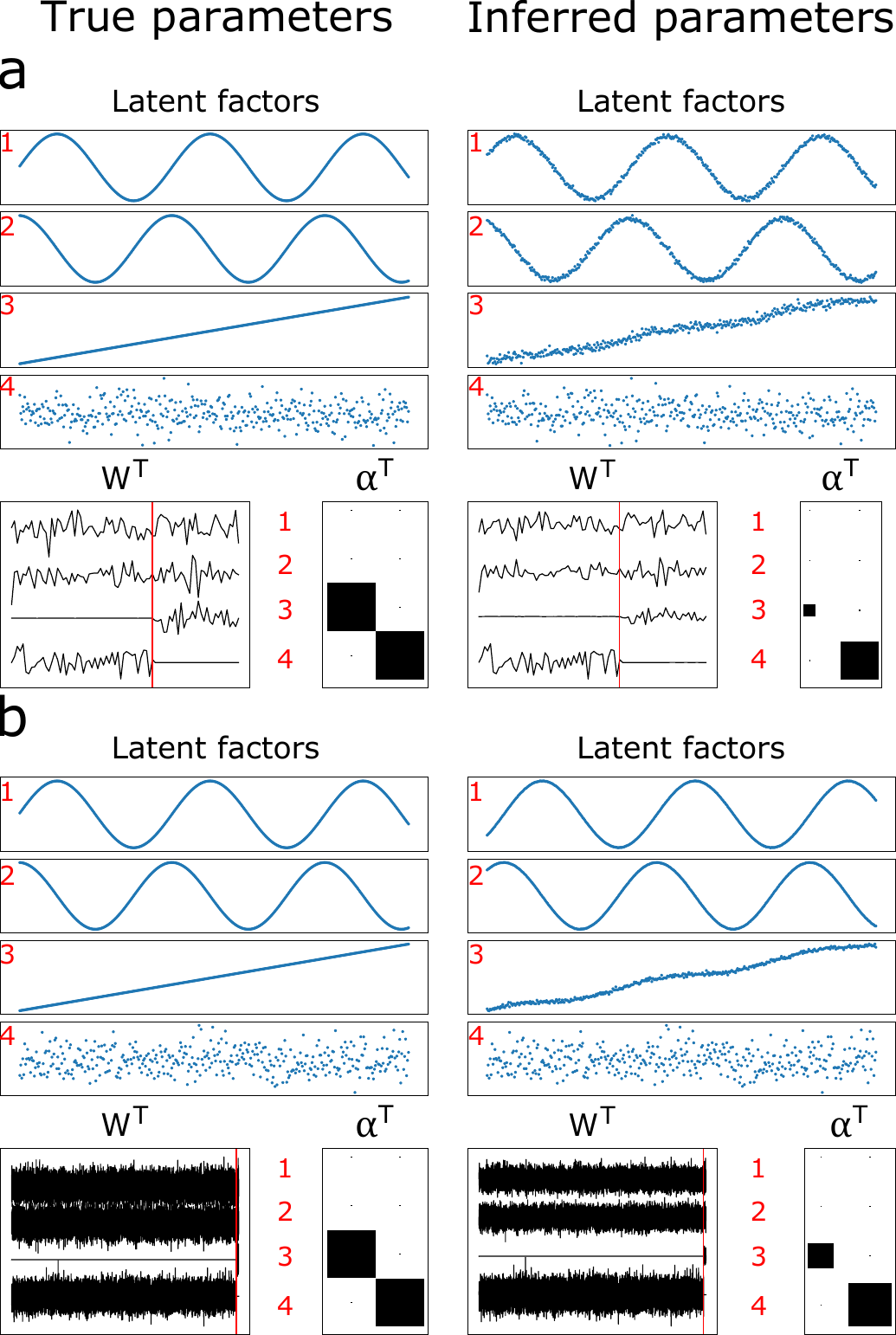}
\caption{True and inferred latent factors and model parameters obtained in the supplementary experiments 1a \textbf{(a)}  and 1b \textbf{(b)}. The latent factors and parameters used to generate the data are plotted on the left-hand side and the ones inferred are plotted on the right-hand side. The four rows on the top represent the 4 latent factors. The loading matrices of the first and second data modality are represented on the left and right-hand side of the red line in $\mathbf{W}^{T}$, respectively. The alphas of the first and second data modality are shown on the first and second column of $\boldsymbol{\alpha}^{T}$, respectively. The small black dots and big black squares represent active and inactive latent factors, respectively.}
\end{figure}

\begin{figure}[H]
\centering
\includegraphics[width=0.85\linewidth]{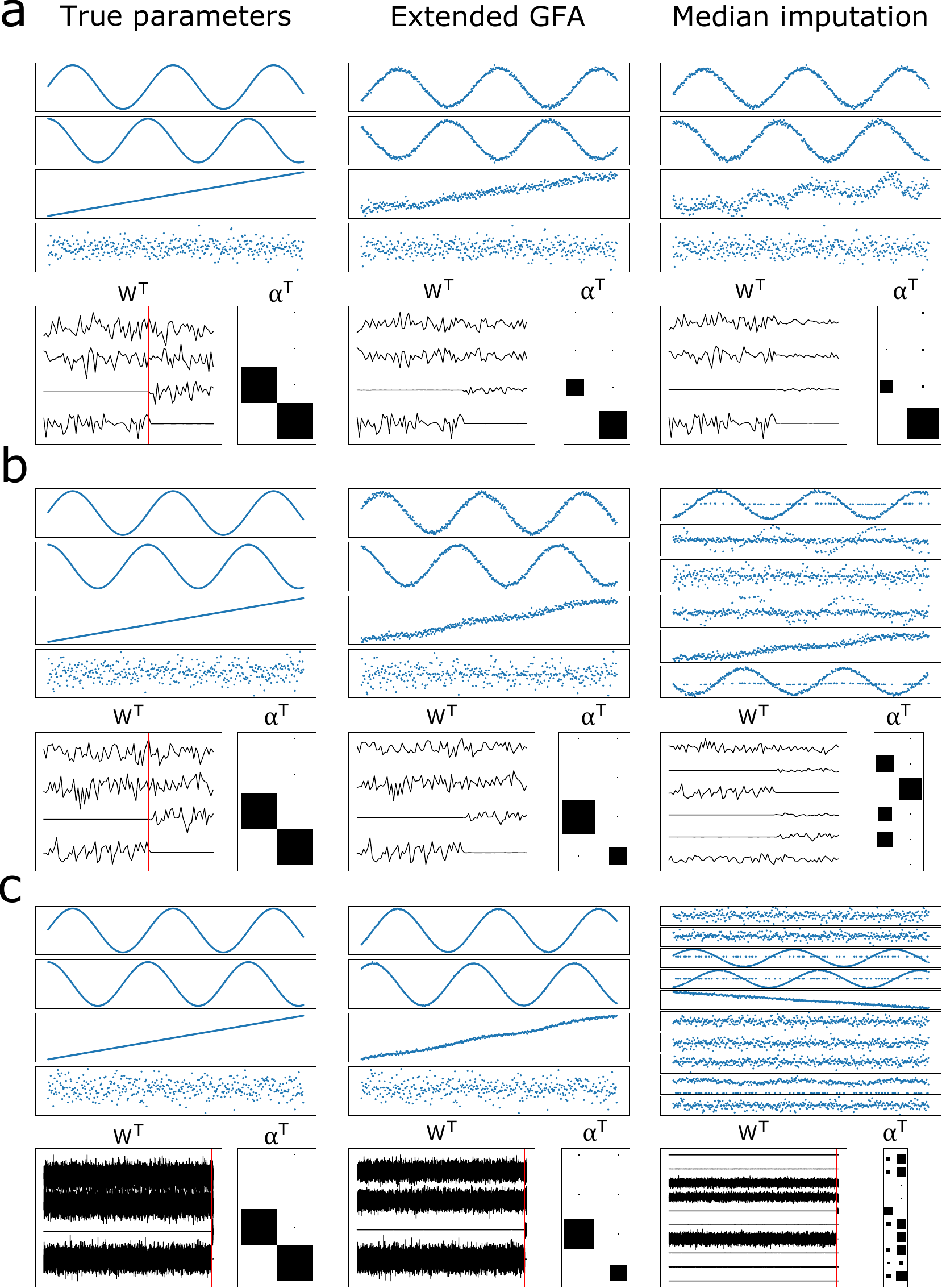}
\caption{True and inferred latent factors and model parameters obtained in the supplementary experiments 2a \textbf{(a)}, 2b \textbf{(b)} and 2c \textbf{(c)}. \textbf{(Left column)} latent factors and model parameters used to generate the data. \textbf{(Middle column)} latent factors and parameters inferred using the proposed GFA extension. \textbf{(Right column)} latent factors and parameters inferred using the median imputation approach. The loading matrices ($\mathbf{W}^{T}$) and alphas ($\boldsymbol{\alpha}^{T}$) can be interpreted as in Supplementary Fig. 1.}
\end{figure}

\begin{table}[H]
    \scriptsize
	\centering
	\renewcommand{\arraystretch}{1.1}
	\renewcommand{\tabcolsep}{2.2pt}
    \caption{Prediction errors of the multi-output prediction tasks obtained in the supplementary experiments 2a-c. The values correspond to the mean and standard deviation of the MSEs across 10 initialisations. The first row shows the MSE between the test observations $\mathbf{X}^{(1)^{\star}}$ and the mean predictions $\langle \mathbf{X}^{(1)^{\star}}|\mathbf{X}^{(2)^{\star}} \rangle$ for all experiments. On the second row, the MSEs between $\mathbf{X}^{(2)^{\star}}$ and $\langle \mathbf{X}^{(2)^{\star}}|\mathbf{X}^{(1)^{\star}} \rangle$ are shown. ours - proposed GFA approach; imputation - median imputation approach; chance - chance level.}
    \begin{tabular*}{\textwidth}{c|ccc|ccc|ccc}
    \toprule
      {} & \multicolumn{3}{c|}{\textbf{Experiment 2a}} &  \multicolumn{3}{c|}{\textbf{Experiment 2b}} & \multicolumn{3}{c}{\textbf{Experiment 2c}} \\
      {} & ours & imputation & chance & ours & imputation & chance & ours & imputation & chance\cr
    \midrule
      $\mathbf{X}^{(1)}|\mathbf{X}^{(2)}$ & \textbf{1.35 $\pm$ 0.34} & 2.87 $\pm$ 1.01 & 2.33 $\pm$ 0.45 & 1.21 $\pm$ 0.10 & 1.16 $\pm$ 0.07 & 2.23 $\pm$ 0.12 & 1.26 $\pm$ 0.06 & 1.17 $\pm$ 0.03 & 2.27 $\pm$ 0.02 \\
      $\mathbf{X}^{(2)}|\mathbf{X}^{(1)}$ & \textbf{0.92 $\pm$ 0.16} & 1.56 $\pm$ 0.28 & 2.04 $\pm$ 0.36 & 0.78 $\pm$ 0.15 & 0.79 $\pm$ 0.15 & 2.34 $\pm$ 0.36 & 0.84 $\pm$ 0.04 & 0.85 $\pm$ 0.04 & 2.35 $\pm$ 0.10 \\
     \bottomrule
    \end{tabular*}
\end{table}

\begin{table}[H]
	\centering
	\renewcommand{\arraystretch}{1.3}
	\renewcommand{\tabcolsep}{8pt}
    \caption{Most relevant shared and modality-specific factors obtained in the complete high dimensional synthetic data (supplementary experiment 1b) according to the proposed criteria. Factors explaining more than $7.5\%$ variance within any data modality were considered most relevant. A factor was considered shared if $0.001 \leq r_{k} \leq 300$, specific to $\mathbf{X}^{(2)}$ if $r_{k} > 300 $ or specific $\mathbf{X}^{(1)}$ if $r_{k} < 0.001$. rvar - relative variance explained; var - variance explained; $r_{k}$ - ratio between the variance explained by $\mathbf{w}_{k}^{(2)}$ and $\mathbf{w}_{k}^{(1)}$.}
    \begin{tabular}{cc|cc|cc|c}
    \toprule
      {} & {} & \multicolumn{2}{c|}{\textbf{rvar} ($\%$)} & \multicolumn{2}{c|}{\textbf{var} ($\%$)} & $\mathbf{r_{k}}$\\
      \multicolumn{2}{c|}{Factors} & $\mathbf{X}^{(1)}$ & $\mathbf{X}^{(2)}$ & $\mathbf{X}^{(1)}$ & $\mathbf{X}^{(2)}$ & var$_{\mathbf{w}_{k}^{(2)}}$/var$_{\mathbf{w}_{k}^{(1)}}$\cr 
    \midrule
      \multirow{2}{*}{\STAB{\rotatebox[origin=c]{90}{\textbf{ \footnotesize Shared}}}} & 1 & 25.09 & 46.03 & 15.39 & 0.19 & 0.01 \\
      & 2 & 25.47 & 35.44 & 15.62 & 0.15 & 9.6 $\times$ 10$^{-3}$ \\
    \midrule
    \multirow{2}{*}{\STAB{\rotatebox[origin=c]{90}{\textbf{ \footnotesize Specific}}}} & 3 & 2.88 $\times$ 10$^{-4}$ & 18.53 & 1.76 $\times$ 10$^{-4}$ & 0.08 & 442.85\\
      & 4 & 49.44 & 8.80 $\times$ 10$^{-5}$ & 30.32 & 3.71 $\times$ 10$^{-7}$ & 1.22 $\times$ 10$^{-8}$  \\
    \bottomrule
    \end{tabular}
\end{table}

\vspace{6pt}
\textit{CCA experiments on synthetic data}
\begin{figure}[H]
\centering
\includegraphics[width=0.9\linewidth]{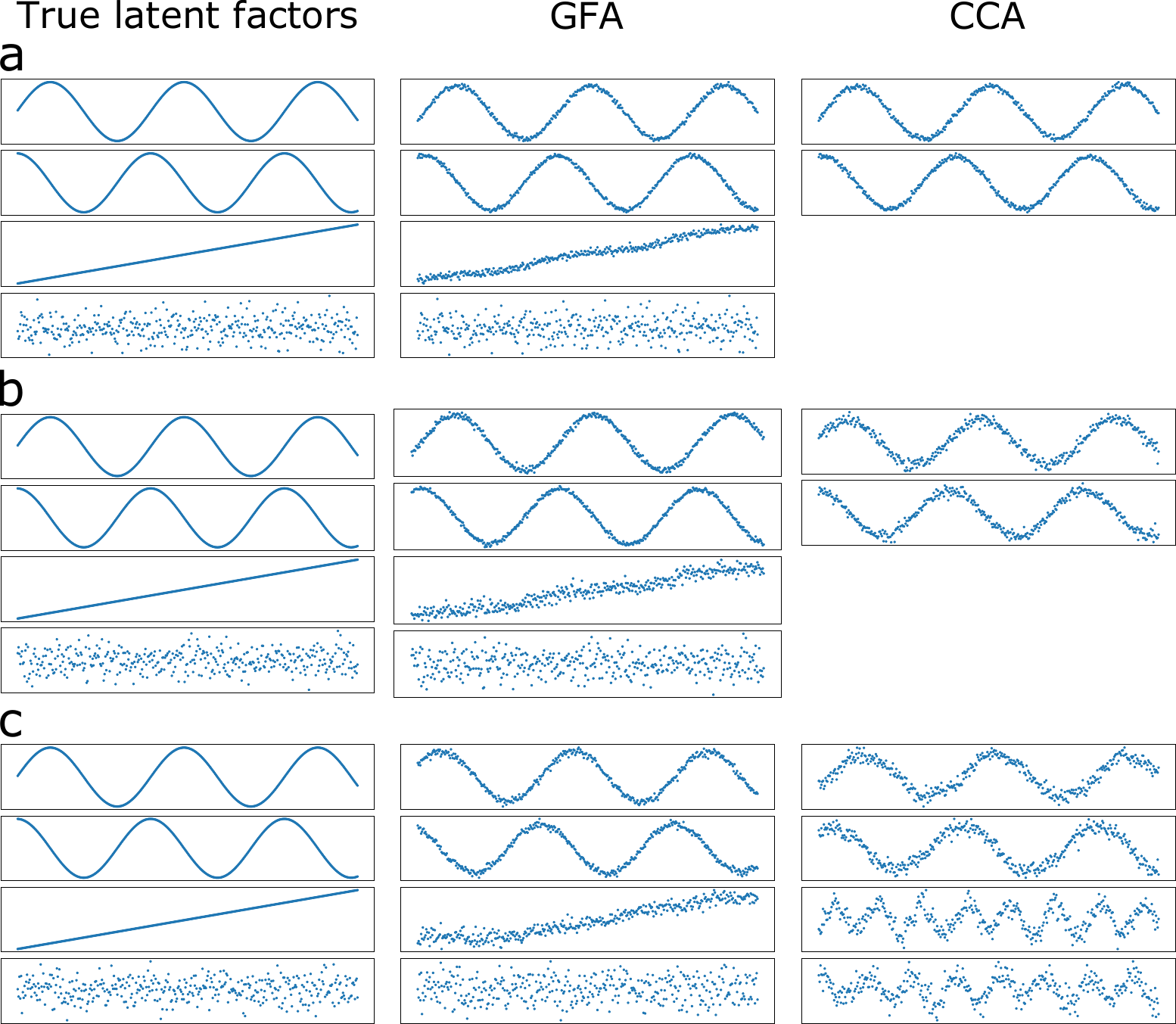}
\caption{True and inferred latent factors model obtained using CCA and GFA. \textbf{(Left column)} latent factors used to generate the data. \textbf{(Middle column)} latent factors inferred using the proposed GFA approach. \textbf{(Right column)} latent factors inferred using CCA. \textbf{(a)} experiment using complete data; \textbf{(b)} experiments using incomplete data, where $20\%$ of the elements of $\mathbf{X}^{(1)}$ and $40\%$ of the elements of $\mathbf{X}^{(2)}$ were randomly removed; \textbf{(c)} experiment using incomplete data, where the elements of $\mathbf{X}^{(1)}$ and $\mathbf{X}^{(2)}$ deviating more than $1\sigma$ (i.e. standard deviation) from the mean were removed.}
\end{figure}

\vspace{6pt}
\textit{GFA experiments on the HCP data}
\vspace{6pt}

\begin{table}[H]
	\centering
	\renewcommand{\arraystretch}{1.2}
	\renewcommand{\tabcolsep}{10pt}
    \caption{Most relevant shared and modality-specific factors obtained in the HCP experiment 2a of the main text ($20\%$ of the elements of the non-imaging matrix missing) according to the proposed criteria. Factors explaining more than $7.5\%$ variance within any data modality were considered most relevant. A factor was considered shared if $0.001 \leq r_{k} \leq 300$, non-imaging (NI) specific if $r_{k} > 300 $ or brain-specific if $r_{k} < 0.001$. rvar - relative variance explained; var - variance explained; $r_{k}$ - ratio between the variance explained by the non-imaging and brain loadings in factor $k$.}
    \begin{tabular}{cc|cc|cc|c}
    \toprule
      {} & {} & \multicolumn{2}{c|}{\textbf{rvar} ($\%$)} & \multicolumn{2}{c|}{\textbf{var} ($\%$)} & $\mathbf{r_{k}}$\\
      \multicolumn{2}{c|}{Factors} & Brain & NI & Brain & NI & var$_{\text{NI}}$/var$_{\text{brain}}$\cr 
    \midrule
      \multirow{4}{*}{\STAB{\rotatebox[origin=c]{90}{\textbf{Shared}}}} & a & 0.159 & 9.44 & 0.012 & 0.028 & 2.42 \\
      & b & 0.065 & 18.152 & 0.005 & 0.005 & 11.32   \\
      & c & 0.036 & 10.539 & 0.003 & 0.031 & 12.04  \\
      & d & 0.015 & 39.330 & 0.001 & 0.117 & 105.10  \\
    \midrule
    \multirow{2}{*}{\STAB{\rotatebox[origin=c]{90}{\textbf{Brain}}}} & a & 13.531 & 6.60 $\times$ 10$^{-5}$ & 0.988 & 1.97 $\times$ 10$^{-7}$ & 1.99 $\times$ 10$^{-7}$  \\
      & b & 12.269 & 0.001 & 0.896 & 4.19 $\times$ 10$^{-6}$ & 4.68 $\times$ 10$^{-6}$  \\
    \bottomrule
    \end{tabular}
\end{table}

\begin{table}[H]
	\centering
	\renewcommand{\arraystretch}{1.2}
	\renewcommand{\tabcolsep}{10pt}
    \caption{Most relevant shared and modality-specific factors obtained in the HCP experiment 2b of the main text ($20\%$ of the subjects missing in the brain connectivity matrix). Factors explaining more than $7.5\%$ variance within any data modality were considered most relevant. A factor was considered shared if $0.001 \leq r_{k} \leq 300$, non-imaging (NI) specific if $r_{k} > 300 $ or brain-specific if $r_{k} < 0.001$. rvar - relative variance explained; var - variance explained; $r_{k}$ - ratio between the variance explained by the non-imaging and brain loadings in factor $k$.}
    \begin{tabular}{cc|cc|cc|c}
    \toprule
      {} & {} & \multicolumn{2}{c|}{\textbf{rvar} ($\%$)} & \multicolumn{2}{c|}{\textbf{var} ($\%$)} & $\mathbf{r_{k}}$\\
      \multicolumn{2}{c|}{Factors} & Brain & NI & Brain & NI & var$_{\text{NI}}$/var$_{\text{brain}}$\cr 
    \midrule
      \multirow{4}{*}{\STAB{\rotatebox[origin=c]{90}{\textbf{Shared}}}} & a & 0.149 & 7.643 & 0.007 & 0.028 & 3.83 \\
      & b & 0.034 & 16.550 & 0.002 & 0.060 & 36.19   \\
      & c & 0.016 & 8.670 & 7.82 $\times$ 10$^{-4}$ & 0.031 & 40.11  \\
      & d & 0.019 & 31.255 & 8.99 $\times$ 10$^{-4}$ & 0.113 & 125.82  \\
    \midrule
    \multirow{2}{*}{\STAB{\rotatebox[origin=c]{90}{\textbf{Brain}}}} & a & 15.625 & 0.0350 & 0.758 & 1.27 $\times$ 10$^{-4}$ & 1.67 $\times$ 10$^{-4}$  \\
      & b & 14.979 & 0.177 & 0.727 & 6.42 $\times$ 10$^{-4}$ & 8.83 $\times$ 10$^{-4}$  \\
    \bottomrule
    \end{tabular}
\end{table}

\begin{figure}[H]
\centering
\includegraphics[width=0.7\linewidth]{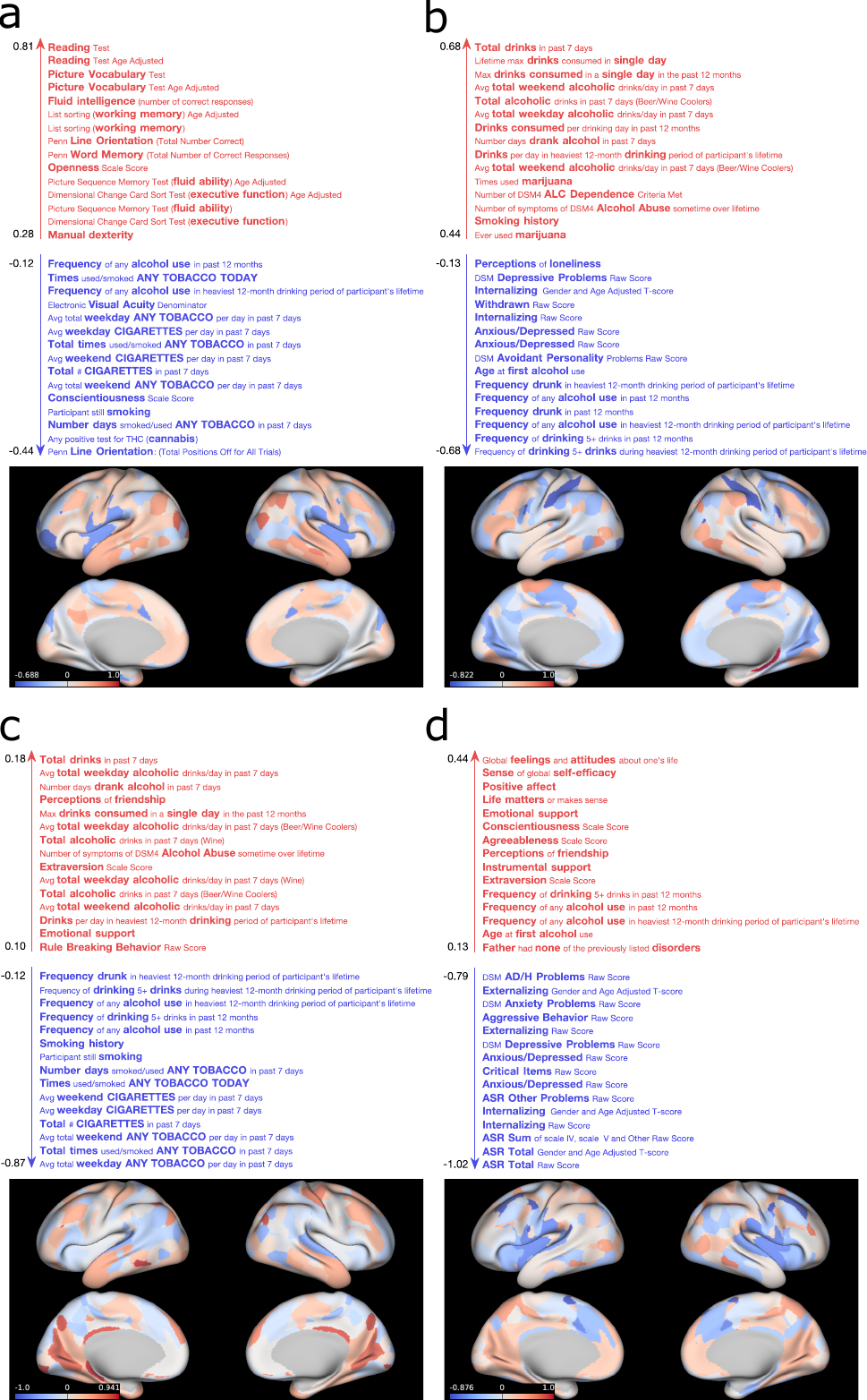}
\caption{Non-imaging measures  and brain networks described by the first \textbf{(a)}, second \textbf{(b)}, third \textbf{(c)} and fourth \textbf{(d)} shared GFA factors obtained in the incomplete data experiment 2a of the main text. For illustrative purposes, the top and bottom 15 non-imaging measures for each factor are shown. The brain surface plots represent maps of brain connection strength increases/decreases, which were obtained as described in the \textit{Surface plots} section.}
\end{figure}

\begin{figure}[H]
\centering
\includegraphics[width=0.74\linewidth]{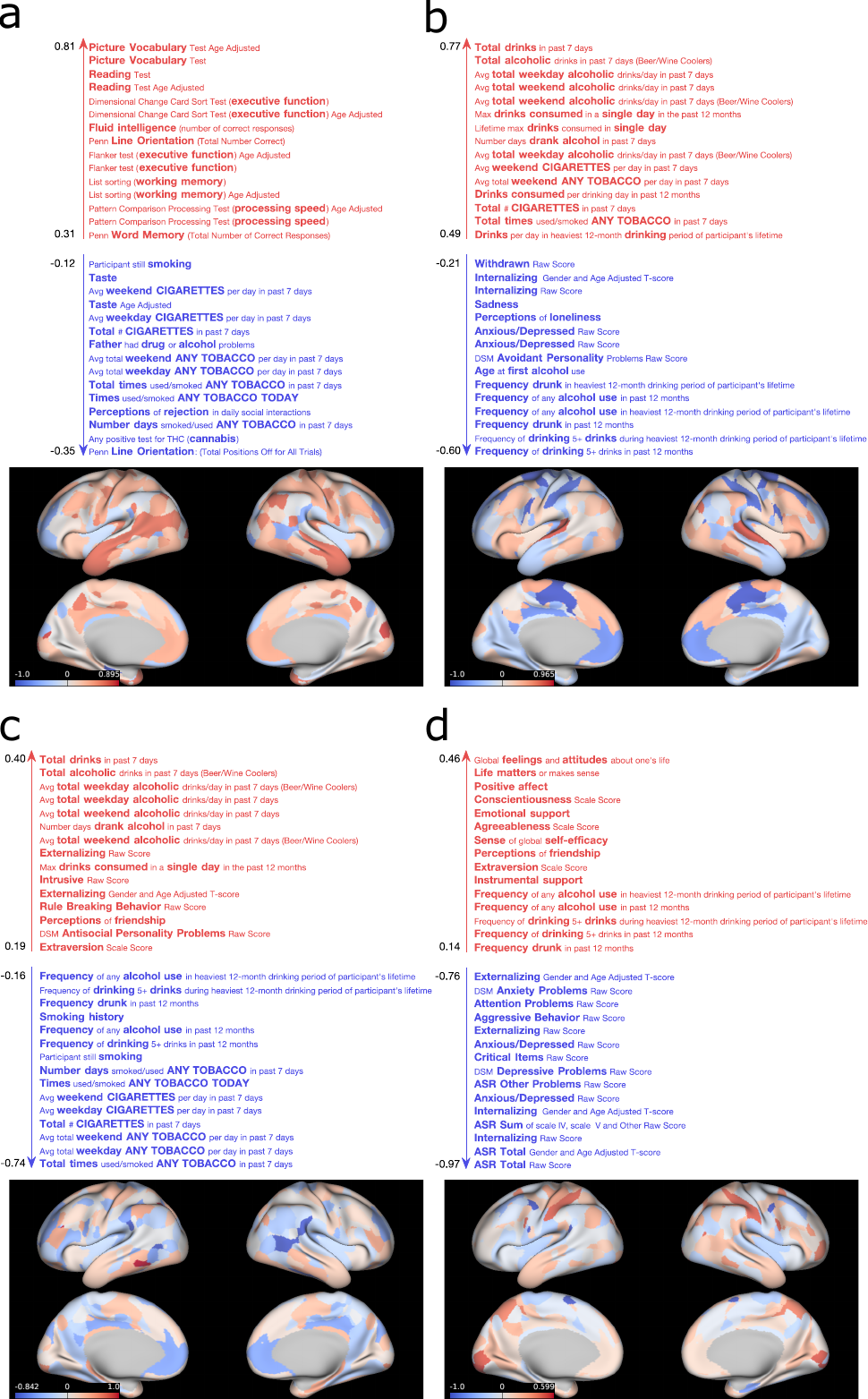}
\caption{Non-imaging measures and brain networks described by the first \textbf{(a)}, second \textbf{(b)}, third \textbf{(c)} and fourth \textbf{(d)} shared GFA factors obtained in the incomplete data experiment 2b of the main text. The top and bottom 15 non-imaging measures for each factor are shown. The brain surface plots represent maps of brain connection strength increases/decreases, which were obtained as described in the \textit{Surface plots} section.}
\end{figure}

\begin{figure}[H]
\centering
\includegraphics[width=\linewidth]{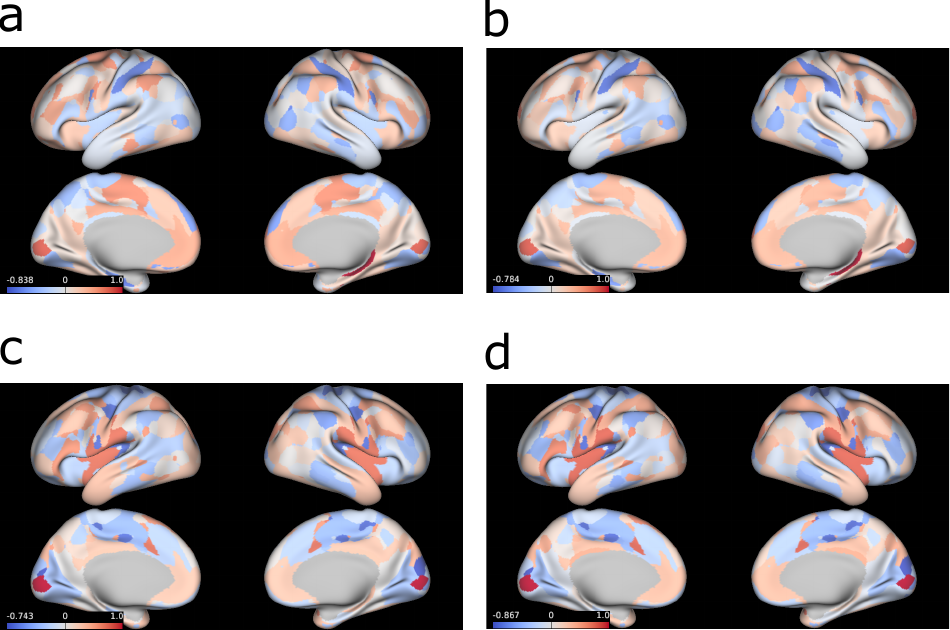}
\caption{Brain networks associated with the brain-specific GFA factors obtained in the incomplete data experiments 2a \textbf{(a,c)}  and 2b \textbf{(b,d)} of the main text. The brain surface plots represent maps of brain connection strength increases/decreases, which were obtained as described in the \textit{Surface plots} section.}
\end{figure}

\begin{figure}[H]
\centering
\includegraphics[width=\linewidth]{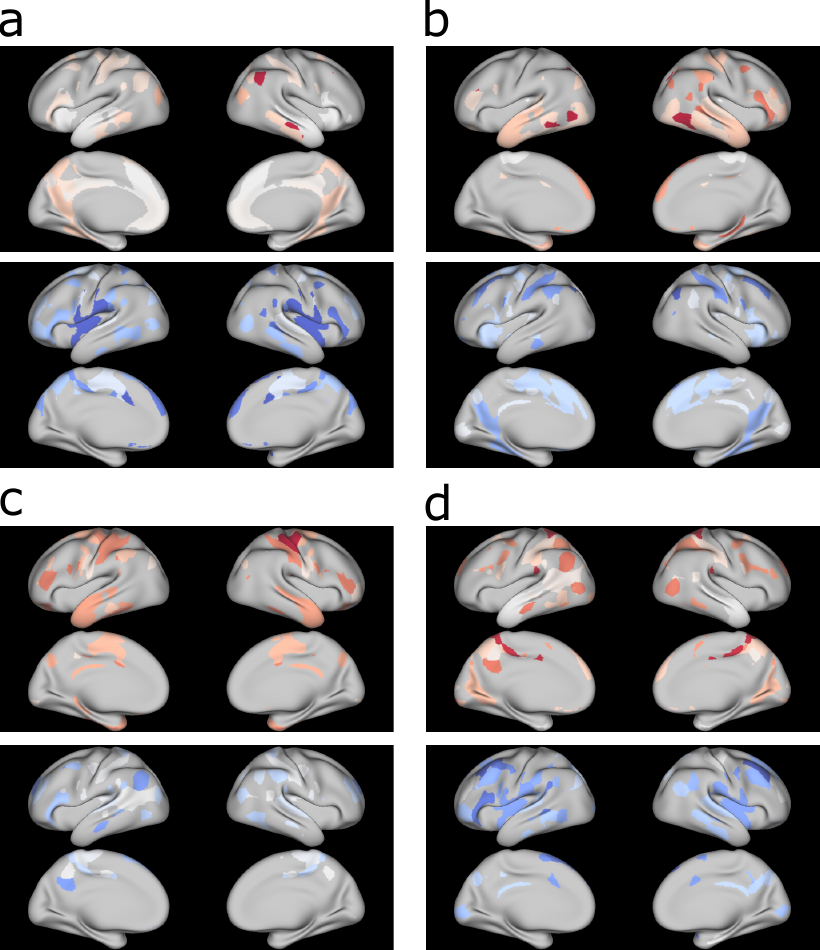}
\caption{Brain surface maps of the brain connection strength increases (red) and decreases (blue) of the first \textbf{(a)}, second \textbf{(b)}, third \textbf{(c)} and fourth \textbf{(d)} shared GFA factors obtained in the HCP experiment with complete data. The distribution of the brain connection strengths was thresholded at the 80th (red) and 20th percentile (blue).}
\end{figure}

\begin{figure}[H]
\centering
\includegraphics[width=0.9\linewidth]{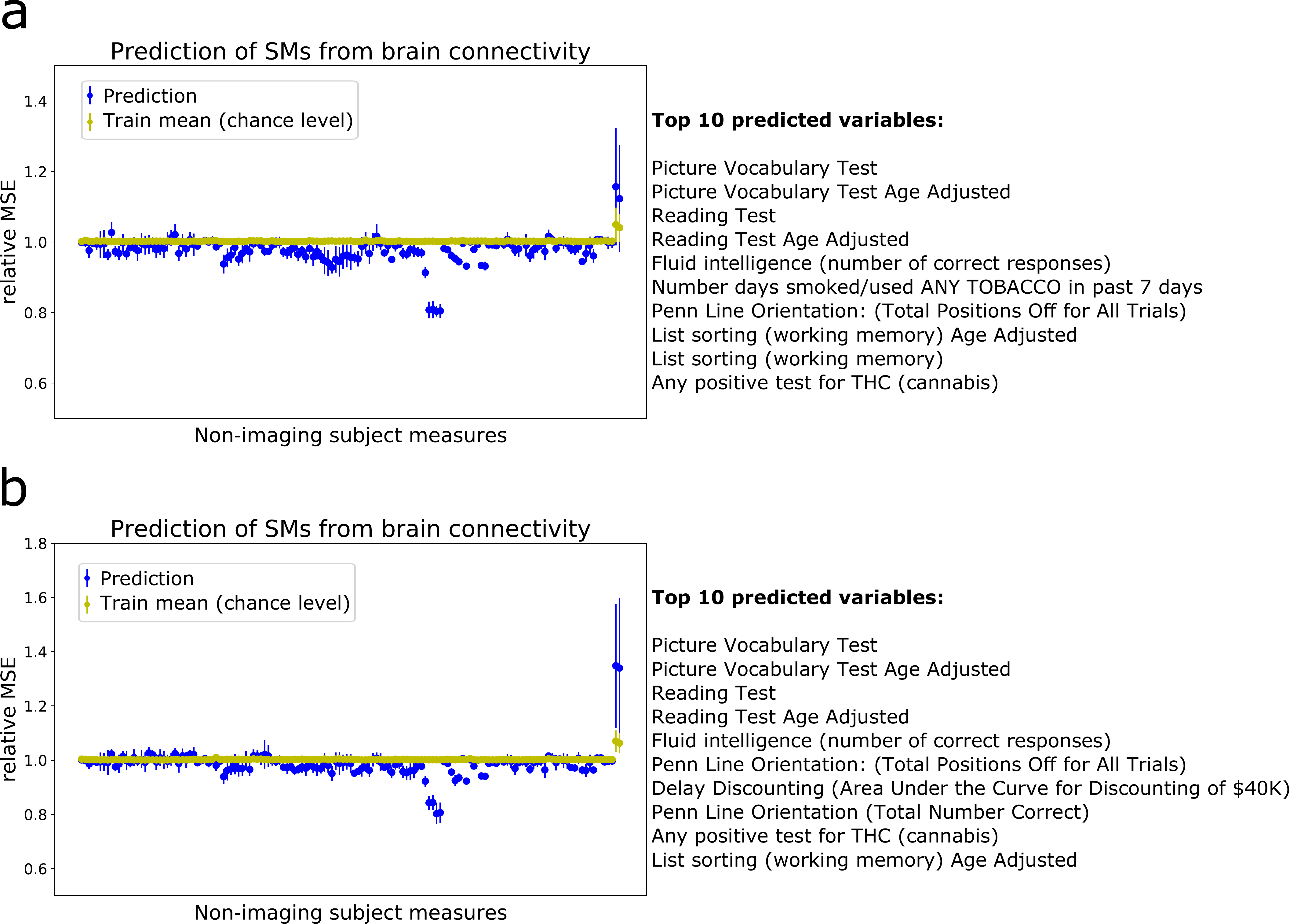}
\caption{Multi-output predictions of the non-imaging measures obtained in the incomplete data HCP experiments 2a \textbf{(a)} and 2b \textbf{(b)} of the main text. The top 10 predicted measures are described on the right. For each non-imaging measure, the mean and standard deviation of the relative MSE between the true and predicted values on the test set was calculated across different random initialisations of the experiments.}
\end{figure}

\textit{CCA experiments on the HCP data}
\vspace{6pt}

To highlight the differences between GFA and CCA, we compared the CCA modes to the GFA factors obtained using complete data (Supplementary Fig. 9 and Supplementary Supplementary Table 5). To interpret the association captured by each CCA mode, we correlated the non-imaging measures and brain connectivity variables with the canonical scores obtained for each data modality (as in \citet{Smith2015}), respectively. The non-imaging measures identified by the first, second and third CCA modes (Supplementary Fig. 9a,b,c) share similarities with the top and bottom non-imaging measures obtained in the first GFA factor (Fig. 6a of the main text). However, the positive brain loadings on the posterolateral and medial default mode networks in the first GFA factor are split between the first and third CCA modes, respectively. The fourth CCA mode is in many respects the inverse of the second GFA factor, both in terms of brain regions (default mode and insular areas) and non-imaging measures, which are dominated by alcohol use (with recent and lifetime use loading at opposite ends of each mode). The fifth CCA mode related most strongly inattention, aggression and antisocial behaviour to positive loadings on posterior insula, and inferior, superior and medial frontal regions. The fourth GFA factor contained these non-imaging measures and low mood/internalising as well. Moreover, the brain loadings in lateral prefrontal and insular cortex were similar across the fifth CCA mode and the fourth GFA factor, as were attention problems and aggression, with conscientiousness loading in the opposite direction.

Finally, the first GFA factor (related to CCA modes 1-3) replicates the findings found by \citet{Smith2015} using CCA applied to approximately 500 subjects (first release of the HCP dataset). Both of these contained loadings related to cognitive performance and tobacco or cannabis use, and brain loadings on default mode areas. Some remaining non-imaging measures in Smith et al’s factor appeared in our fourth GFA factor (related to life satisfaction and aggression), which strongly related to different forms of psychopathology.

\begin{table}[H]
	\centering
	\renewcommand{\arraystretch}{1.2}
	\renewcommand{\tabcolsep}{10pt}
    \caption{Pearson's correlations between the most relevant GFA factors (in the complete data experiment described in the main text) and the CCA modes (Supplementary Fig. 9) obtained by applying CCA as described in the \textit{CCA experiments on the HCP data} section. The values in bold represent the highest absolute correlations between a given CCA mode and the GFA factors.}
    \begin{tabular}{cc|cccc|cc}
    \toprule
      \multicolumn{2}{c|}{} & \multicolumn{6}{c}{\textbf{GFA}}\\
      \multicolumn{2}{c|}{} & \multicolumn{4}{c}{\textbf{Shared}} & \multicolumn{2}{|c}{\textbf{Brain-specific}}\\
      \multicolumn{2}{c|}{} & a & b & c & d & a & b\cr 
    \midrule
      \multirow{4}{*}{\STAB{\rotatebox[origin=c]{90}{\textbf{CCA}}}} & a & \textbf{0.605} & 0.011 & 0.105 & 0.064 & 0.093 & 0.347 \\
      & b & \textbf{0.380} & 0.112 & 0.050 & 0.190 & 0.093 & 0.081  \\
      & c & 0.231 & 0.112 & 0.206 & 0.065 & \textbf{0.299} & 0.048 \\
      & d & 0.009 & \textbf{0.191} & 0.039 & 0.061 & 0.083 & 0.036 \\
      & e & 0.052 & 0.092 & 0.115 & 0.173 & 0.031 & \textbf{0.386} \\
    \bottomrule
    \end{tabular}
\end{table}

\begin{figure}[H]
\centering
\includegraphics[width=0.95\linewidth]{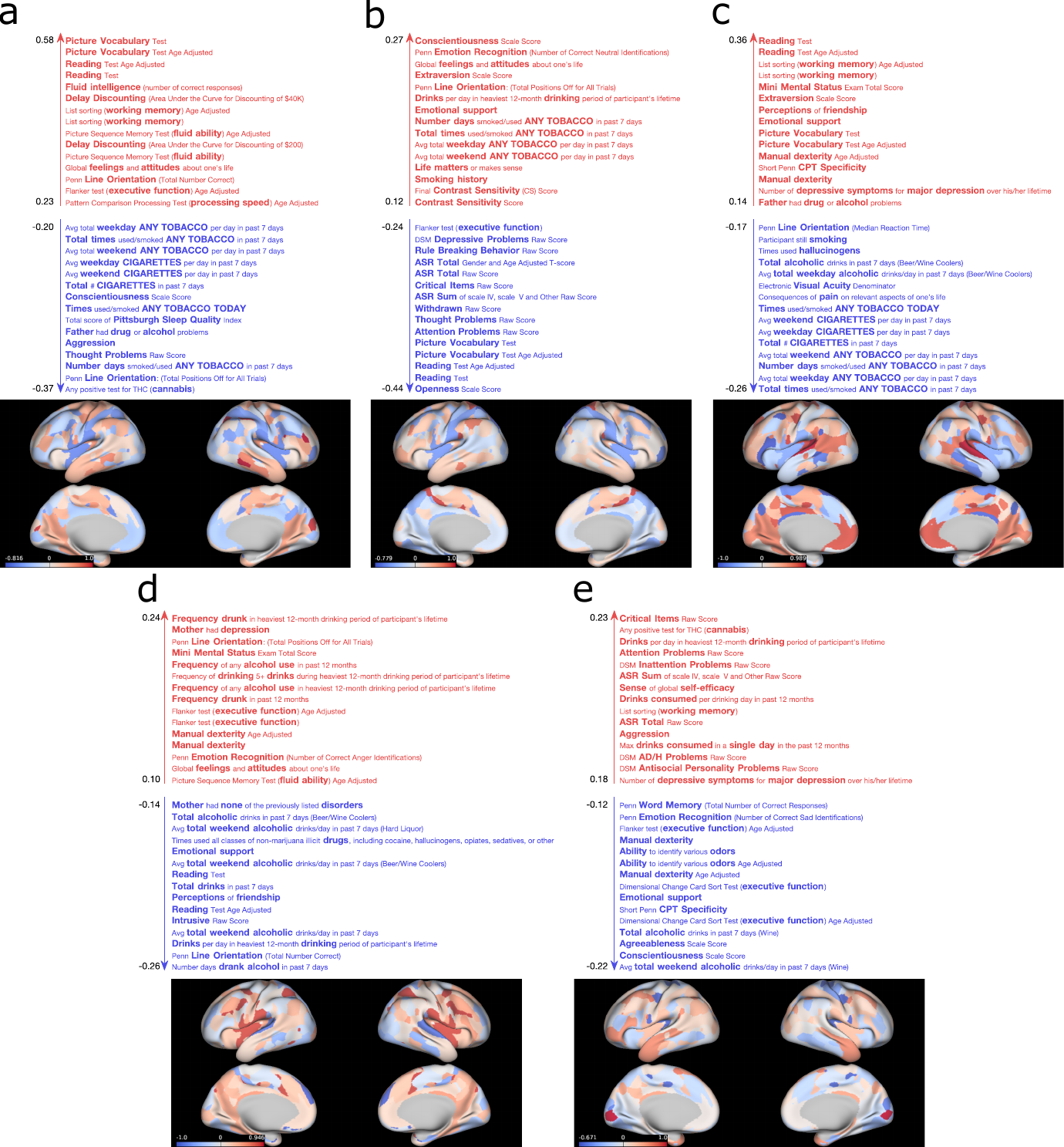}
\caption{Non-imaging measures and brain networks correlated with the CCA modes obtained by applying CCA as described in the \textit{CCA experiments on the HCP data} section. For illustrative purposes, the top and bottom 15 non-imaging measures for each factor are shown. The brain surface plots represent maps of brain connection strength increases/decreases, which were obtained as described in the \textit{Surface plots} section.}
\end{figure}

\textit{Alcohol use loadings of the second GFA factor}
\vspace{6pt}

The second GFA factor (Fig. 6b of the main text) has puzzlingly opposing loadings of frequency of alcohol use versus total alcohol drunk in the last seven days. This is probably because the distributions of ``total amount" answers are very skewed, with most subjects reporting zero, hence a lot of variance can be explained by this rather paradoxical set of loadings (Supplementary Fig. 10). Alternatively, it might be that these alcohol use items represent two different behaviours, where ``total amount" answers are related to a more short-term alcohol use and the ``frequency" answers might represent more long-term and consistent alcohol use.

\begin{figure}[H]
\centering
\includegraphics[width=\linewidth]{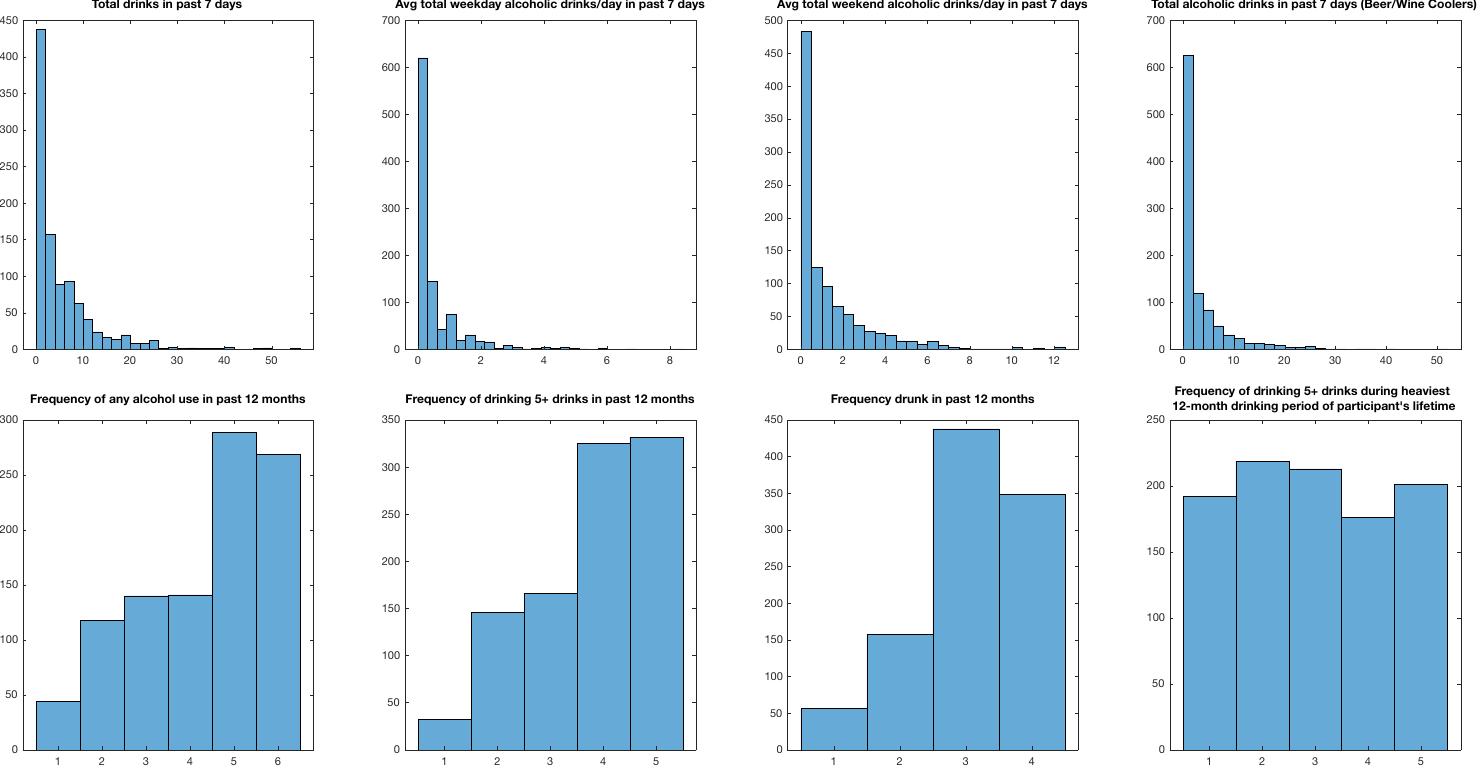}
\caption{Histograms of the top 4 variables \textbf{(top)} and bottom 4 variables \textbf{(bottom)} of the second shared GFA factor displayed in Fig. 6b of the main text.}
\end{figure}
\end{document}